\documentclass[a4paper,12pt]{article}

\usepackage{amsmath, amssymb, amsthm, amsfonts, amscd, epsfig, graphicx, color}
\definecolor{extlinkz}{rgb}{0,0,1}
\definecolor{intlinkz}{rgb}{1,0,0}
\definecolor{citlinkz}{rgb}{0,.6,.1}
\usepackage[pdfpagelabels,plainpages=false,colorlinks=true,citecolor=citlinkz,linkcolor=intlinkz,urlcolor=extlinkz,urlbordercolor={0 0 0}]{hyperref}

\newcommand{\R}{\mathbb R}
\newcommand{\C}{\mathbb C}
\newcommand{\N}{\mathbb N}
\newcommand{\T}{\mathcal T}
\newcommand{\M}{\mathcal M}
\newcommand{\Ll}{\mathcal L}

\newcommand{\matr}[4]{
\left(
\begin{array}{cc}
#1 & #2\\
#3 & #4
\end{array}
\right)
}
\newcommand{\Id}[1]{\mathbb{I}_{#1 \times #1}}
\newcommand{\ol}{\overline}

\title{Cortical spatio-temporal dimensionality reduction\\ for visual grouping}

\author{Giacomo Cocci, Davide Barbieri, Giovanna Citti, Alessandro Sarti}

\begin{document}

\maketitle

\vspace{-20pt}
\abstract{\noindent The visual systems of many mammals, including humans, is able to integrate the geometric information of visual stimuli and to perform cognitive tasks already at the first stages of the cortical processing. This is thought to be the result of a combination of mechanisms, which include feature extraction at single cell level and geometric processing by means of cells connectivity. We present a geometric model of such connectivities in the space of detected features associated to spatio-temporal visual stimuli, and show how they can be used to obtain low-level object segmentation. The main idea is that of defining a spectral clustering procedure with anisotropic affinities over datasets consisting of embeddings of the visual stimuli into higher dimensional spaces. Neural plausibility of the proposed arguments will be discussed.}

{
\hypersetup{linkcolor=black}
\setcounter{tocdepth}{3}
\tableofcontents
}

\section{Introduction}

It is well understood from the psychological theory of the Berliner Gestalt that local properties of the visual stimulus, such as neighboring, good continuation and common fate have a central role in the execution of global visual tasks like image segmentation and grouping \cite{GestaltW,Gestalt}.

A key concept for the understanding of visual perceptual tasks is that of \emph{association fields}, that was introduced in \cite{Field1993} to describe the sructure of the field of good continuation, underlying the recognition of perceptual units in the visual space. These results have been obtained by psychophysical experiments, presenting stimuli made of an ensemble of oriented patches, a subset of which was consistently aligned along a continuous path. The study of these phenomena allowed to identify the properties that a stimulus should have near a given patch in order to recognize such sample as belonging to a curvilinear path, namely co-linearity and co-circularity. The detection of these properties is indeed compatible with the functional behavior of simple cells in the primary visual cortex V1 as linear feature detectors for local orientations \cite{Hubel1977}.

Several physiological experiments showed how the principles of association fields seem to be implemented in the V1 of mammals, where long-range horizontal connections preferentially link columns of neurons having similar preferred orientation \cite{Bosking1997}. On the other hand, by interpreting cortical columns as directional differential operators, in \cite{CS} it is shown how this specialized functional organization of V1 naturally leads to a geometric model of the association fields. The field lines are modeled with a family of integral curves on a contact structure based on the Lie algebra of the group of rigid motions of the Euclidean plane $SE(2)$. This geometric approach lies within a well established research line founded by the seminal works \cite{Koenderink, Hoffman, Mumford, Petitot1999}, whose current state of the art can be found in \cite{CSbook} and references therein.

Further phenomenological experiments demonstrated the central role for the perception of global shapes of the features of movement direction and velocity \cite{Rainville2005} and that, similarly to what happens for the integration of spatial visual information, the brain is capable to predict complex stimulus trajectories, and to group together elements having similar motion or apparent motion paths \cite{Verghese1999, Verghese2000, Watamaniuk2005}. Also in this case, the detection of such features is performed at the level of V1, where specialized cells show spatio-temporal behaviors optimized for the detection of local velocities \cite{DeAngelis1995, CBS}.

The analysis of spatio-temporal properties and organization of cortical visual neurons, together with the indications given by experimental results on visual  spatial and motion integration, have recently led to extensions of the $SE(2)$ model which include local stimulus velocity. In \cite{BCCS}, new classes of spatio-temporal connectivities were indeed introduced, providing a geometric model of association fields in the five dimensional contact structure of cells positions and activation times, together with the locally detected features of orientations and velocities. Such a structure embeds the purely spatial geometry in a layered fashion, and coherently integrates the association mechanisms as extensions to a higher dimensional space. Their capabilities in the elaboration of trajectories, and in the tasks of spatio-temporal image completion, actually showed good accordance with previous psychophysical and physiological results.

The aim of the present paper is to study the capabilities of such geometric spatio-temporal connectivities with respect to the tasks of image segmentation and grouping. Such tasks are addressed with spectral analysis, and in particular we will discuss and refine the proposed connectivity structures in order to properly describe them as operators on high dimensional feature spaces. The approach we will follow is to represent a spatio-temporal stimulus as a dataset in a feature space, and consider it as a weighted graph whose affinity matrix is determined by the geometric connectivity. The grouping mechanisms that we will describe arise from a spectral clustering of the associated graph Laplacian, making use of the probabilistic framework introduced in \cite{meilashi} followed by an adaptation of the simple and robust clustering technique proposed in \cite{Kannan} and, when dealing with nonsymmetric affinities, we followed the ideas introduced in \cite{Pentney2005}. We have tried to stuck with minimal hypotheses at the algorithmic level in order to keep focus on the role of the kernels. Our main results will show that the introduced spatio-temporal geometry provides connectivities suitable to robustly group spatio-temporal stimuli, but also that a connectivity pattern based also on local stimulus velocity can enhance the spatial grouping capabilities of a visual system.

With respect to actual neural implementations of such principles, apart from the study in \cite{BCCS}, we can see that psychophysical experiments such as those of \cite{Geisler2001, Hess2003, Ledgeway2005} addressed the problems of existence of association fields for local directions of motion and their role in visual grouping, and of comparison of human grouping with co-circular correlations in natural image statistics. Moreover, recent results of \cite{SartiCittiPercepts} show how spectral analysis of the spatial connectivity introduced in \cite{CS}, that is the more basic one studied in the present paper, is actually implemented by the neural population dynamics of primary visual cortex.
This suggests a fundamental role of spectral mechanisms in the phenomenology of perception, indicating that they may be concretely performed by the visual system, and hence providing a stronger motivation for the present detailed spectral analysis of connectivities and their segmentation properties of visual stimuli.

Finally, we would like to recall also that while co-circularity is naturally implemented in the presently studied kernels (see e.g. \cite{SCS}), its use for the definition of affinity matrices whose spectra could be suited for line grouping was suggested already in \cite{Perona1998}.


The plan of the paper is the following. We will start in Section \ref{sec:geometry} with a description of the geometry arising from the spatio-temporal functional architecture of the visual cortex, as introduced in \cite{BCS2013,BCCS}, providing more detailed arguments on how to construct and implement the connectivity kernels we will use.
In Section \ref{sec:cdm} we will recall the needed notions of spectral analysis on graphs, and introduce the basic clustering principles that we will adopt. Then we will show how to construct affinity matrices based on the spatio-temporal cortical geometry, providing specific discussion on how to deal with the different kinds of asymmetries present in the connectivities.
Or main results will constitute Section \ref{sec:grouping}, where we will use the previously introduced spectral clustering algorithm to automatically extract perceptual information from artificial stimuli living in the cortical feature spaces described in Section \ref{sec:geometry}. In particular, we will propose two different connectivity mechanisms to perform spatio-temporal segmentation of motion contours and shapes, also providing parametric evaluations of the kernel performances, and we will discuss the relations with neural processes studied in several psychophysiological experiments.

\paragraph{Acknowledgements}
D. Barbieri was supported by a Marie Curie Intra European Fellowship (Prop. N. 626055) within the 7th European Community Framework Programme.

\section{The geometry of V1}
\label{sec:geometry}

In this section we present a model of the functional architecture of the visual cortex as a contact structure, where the cortical connectivity is implemented as a diffusion process along its admissible directions. This same approach was taken in \cite{BCCS}, where it was compared with psychophysical and physiological behaviors of the visual system.

\subsection{The cortical feature space}

It is well known since the fundamental studies of Hubel and Wiesel \cite{Hubel1977, Hubel} that primary visual cortex (V1) is one of the first physiological layers along the visual pathway to carry out geometrical measurements on the visual stimulus, decomposing it in a series of local feature components. The development of suitable electrophysiological techniques \cite{Ringach2004} has made possible to reconstruct the linear filtering behavior of V1 simple and complex cells, i.e. their spatio-temporal receptive profiles (RPs).

\begin{figure}
\centering
\includegraphics[height=.4\textwidth, width=.4\textwidth]{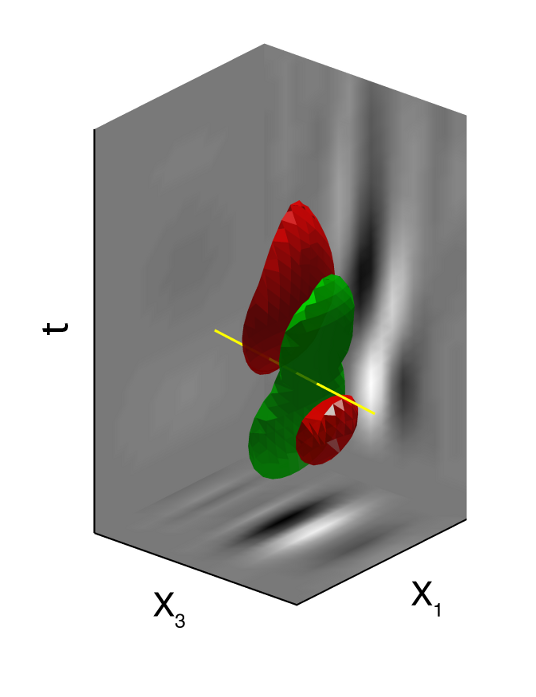}
\includegraphics[width=.52\textwidth]{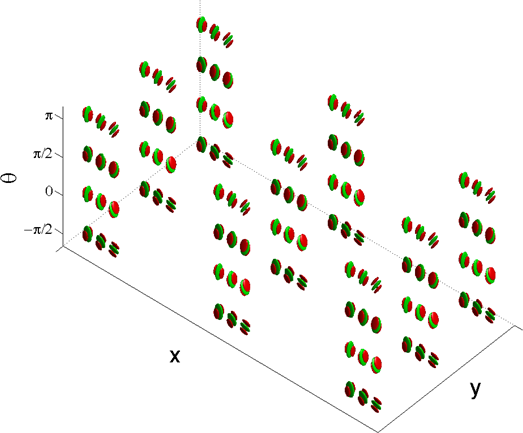}
\caption{Left: isolevel surfaces of a V1 inseparable receptive profile depicted in space-time. Green and red surfaces enclose, respectively, excitatory and inhibitory regions. The yellow line indicates the direction of the local vector $\vec{X}_{\theta,v}$ (see text for details). Right: schematization of the feature-wise organization of the primary visual cortex. For each spatio-temporal point $(x,y,t)$ of the image hyperplane there is a two dimensional fiber of representing local orientation $\theta$ and local velocity $v$.}\label{fig:features}
\end{figure}

The RPs of orientation-selective cells in V1 have classically been modeled with two-dimensional Gabor functions \cite{Jones1987b}, which basically compute a local approximation of the directional derivative of the visual stimulus, minimizing the uncertainty between localization in position and spatial frequency \cite{Daugman}. On the other hand, spatio-temporal RPs of V1 simple cells can be modeled by 3-dimensional Gabor functions of the form \cite{CBS}
\begin{equation}\label{eq:RP}
g^\sigma_{q,p}(x,y,t) = e^{ 2 \pi i ( p_x ( x - q_x ) + p_y ( y - q_y ) + p_t ( t - q_t ) ) } e^{ - \left(\frac{(x-q_x)^2}{2 \sigma_x^2} + \frac{(y-q_y)^2}{2 \sigma_y^2} + \frac{(t-q_t)^2}{2 \sigma_t^2} \right)} ,
\end{equation}
where $q=(q_x,q_y,q_t)$ is the spatio-temporal center of the Gabor filter, $p=(p_x, p_y, p_t)$ is the spatio-temporal frequency and $\sigma=(\sigma_x,\sigma_y,\sigma_t)$ is the spatio-temporal width. One of the crucial features of (\ref{eq:RP}) is minimization of the uncertainty of simultaneous measurements in space-time and frequency. It is worth noting that this model strictly captures the features of so-called inseparable RPs, tuned for position, orientation and motion detection, depicted in Fig. \ref{fig:features} left. On the other hand, separable RPs can be obtained as linear superpositions.

Further analyses have also shown that the Gabor parameter distribution found in cortical cells cover only a subset of the whole Gabor family. Such subsets are optimized for the detection of the local features of orientation $\theta$ and speed $v$ \cite{CBS,BCS2013}
\begin{displaymath}
\begin{array}{rcl}
\theta & = & \arctan \left( \frac{p_y}{p_x} \right)\\
v & = & \frac{p_t}{\sqrt{ p_x^2 + p_y^2 }}
\end{array}
\end{displaymath}
that can be interpreted as fundamental features of the visual stimulus. For this reason we will not deal with the dependence on the spatial frequency $\kappa = \sqrt{p_x^2 + p_y^2}$ or on the scales $\sigma$, but consider RPs of the form
\begin{equation}\label{eq:RPupto}
g_{q,\theta}^{\sigma,\kappa}(x,y,t) = e^{ - 2 \pi i \kappa ( ( x - q_x )\cos\theta + ( y - q_y )\sin\theta + v ( t - q_t ) ) } e^{ - \left(\frac{(x-q_x)^2}{2 \sigma_x^2} + \frac{(y-q_y)^2}{2 \sigma_y^2} + \frac{(t-q_t)^2}{2 \sigma_t^2} \right)}
\end{equation}
where $\kappa$ and $\sigma$ are considered as fixed.
This corresponds to a neural processing stage where the visual stimulus is lifted from the spatio-temporal image space $\R^2\times\R^+$ to the extended 5-dimensional feature space
$$
\M_T = \R^2\times\R^+\times S^1\times\R^+ = \left\{ \eta = (q_x,q_y,q_t,\theta,v) \right\} ,
$$
where every point $\eta \in \M_T$ corresponds to a filter (\ref{eq:RPupto}), as in Fig. \ref{fig:features} right. The activity of V1 simple cells is then considered as modeled by the map
\begin{equation}\label{eq:filtering}
f \mapsto F^{\sigma,\kappa}(q,\theta,v) \doteq \langle g^{\sigma,\kappa}_{q,\theta}, f\rangle_{L^2(\R^3)} .
\end{equation}

\subsection{Connectivity as a differential constraint}

The functional behavior of V1 simple cells modeled by (\ref{eq:filtering}) can be interpreted as a finite scale spatio-temporal directional derivative of the stimulus around position $q$, performed along the direction
\begin{displaymath}
\vec{X}_{\theta,v} = (\cos\theta, \sin\theta,-v)
\end{displaymath}
expressed in the coordinates $\{\hat{e}_{x},\hat{e}_{y},\hat{e}_{t}\}$. Accordingly, this derivation is maximal along the direction of the gradient of $f$.
This implies that the lifting to $\M_T$ of any smooth level set of $f$ is always orthogonal to the vector field
$$
\vec{X} = (\cos\theta, \sin\theta,-v,0,0) \in \T_\eta\M_T
$$
expressed in the coordinates $\{\hat{e}_{q_x},\hat{e}_{q_y},\hat{e}_{q_t},\hat{e}_{\theta},\hat{e}_{v}\}$, where $\T_\eta\M_T$ stands for the tangent space of $\M_T$ at $\eta = (q_x,q_y,q_t,\theta,v)$.

Hence the present problem induces to consider as admissible surfaces on $\M_T$ those whose tangent space at any point is spanned by the vector fields
\begin{equation}\label{eq:vectorfields}
\begin{array}{l}
\vec{X}_1 = ( -\sin\theta,  \cos\theta, 0, 0, 0) \, , \ \vec{X}_2 = (0,0,0,1,0)\\
\vec{X}_4 = (0,0,0,0,1) \, , \ \vec{X}_5 = ( v \cos\theta, v \sin\theta, 1,0,0)
\end{array}
\end{equation}
defining the orthogonal complement to $\vec{X}$ in $\T\M_T$.
\newpage
The four dimensional hyperplanes generated by $\{X_1,X_2,X_4,X_5\}$ is called contact plane, and the whole structure is named a contact structure. Contact structures have been used for modeling the functional architecture of the visual cortex in several works, see e.g. \cite{Petitot1999,CS}.

Due to this contact structure,
the connectivity among V1 cells on $\M_T$ can be modeled geometrically as in \cite{BCCS} in terms of advection-diffusion processes along the directions of the vector fields (\ref{eq:vectorfields}). Two corresponding stochastic processes were introduced in order to provide concrete realizations of the mechanisms of propagation of information along connections.

A first mechanism, aimed to model connectivity along lifted contours of a spatial image at a fixed time, consists of a propagation along the direction $\vec{X}_1$ forced by a diffusion over $\vec{X}_2$ and $\vec{X}_4$. This will be used for a single-frame segmentation out of a spatio-temporal streaming. It lives on a codimension 1 submanifold of $\M_T$ at fixed time $t_0$ that we will call
$\M_0 = \R^2\times S^1\times\R^+ = \left\{ \xi = (q_x,q_y,q_t,\theta,v) : q_t = t_0\right\}$,
and can be formally described by the following system of stochastic differential equations
\begin{equation}\label{eq:pathV}
\left\{
\begin{array}{rcl}
dq_x(s) & = & -\sin\theta(s)\, ds\vspace{4pt}\\
dq_y(s) & = & \cos\theta(s)\, ds\vspace{4pt}\\
d\theta(s) & = & \kappa \ dW_1(s) \vspace{4pt}\\
dv(s) & = & \alpha \ dW_2(s) \vspace{6pt}\\
\xi(s=0) & = & \xi_0
\end{array}
\right.
\end{equation}
where $W = (W_1, W_2)$ is a two dimensional Brownian motion and $(\kappa,\alpha)$ are the corresponding diffusion constants. The Fokker-Plank equation associated to this process, which provides a transition probability density $\rho_0$, is
\begin{equation}\label{eq:FPV}
(\partial_s - L_0) \rho_0(\xi,s|\xi_0,0) = \delta(\xi - \xi_0)\delta(s)
\end{equation}
where the evolution operator is given by $L_0 = X_1 - \kappa^2 X_2^2 - \alpha^2 X_4^2$ and, as customary, we have denoted with $X_i$ the directional derivative along the vector field $\vec{X}_i$.

A second mechanism, aimed to model connectivity among moving contours of a spatio-temporal stimulus, will be used for spatio-temporal segmentation of apparent point trajectories. It consists of a propagation along $\vec{X}_5$ again forced by a diffusion over $\vec{X}_2$ and $\vec{X}_4$, and is described by the stochastic process on $\M_T$
\begin{equation}\label{eq:pathT}
\left\{
\begin{array}{rcl}
dq_x(s) & = & v\cos\theta(s)\, ds\vspace{4pt}\\
dq_y(s) & = & v\sin\theta(s)\, ds\vspace{4pt}\\
dq_t(s) & = & ds\vspace{4pt}\\
d\theta(s) & = & \kappa \ dW_1(s) \vspace{4pt}\\
dv(s) & = & \alpha \ dW_2(s) \vspace{4pt}\\
\eta(s=0) & = & \eta_0 .
\end{array}
\right.
\end{equation}
The Fokker-Plank equation associated to this process, which provides a transition probability density $\rho_T$, is
\begin{equation}\label{eq:FPT}
(\partial_s - L) \rho_T(\eta,s|\eta_0,0) = \delta(\eta - \eta_0)\delta(s)
\end{equation}
where the evolution operator is given by $L = X_5 - \kappa^2 X_2^2 - \alpha^2 X_4^2$.

The structure of these processes explicitly assigns different roles to the spatio-temporal variables $q$, where the stimulus is defined, and to the engrafted variables $(\theta,v)$. More precisely, we have advection in the $q$ variables while diffusion occurs in the $(\theta,v)$ variables. It is worth noting that this construction naturally extend the process proposed by Mumford in \cite{Mumford} for the case of static images, which consists of a propagation along the direction $\vec{X}_1$ forced by a diffusion over $\vec{X}_2$, on the three dimensional submanifold $\M_3 = \R^2 \times S^1 = \{\zeta = (q_x,q_y,\theta)\}$
\begin{equation}\label{eq:path3}
\left\{
\begin{array}{rcl}
dq_x(s) & = & -\sin\theta(s)\, ds\vspace{4pt}\\
dq_y(s) & = & \cos\theta(s)\, ds\vspace{4pt}\\
d\theta(s) & = & \kappa \ dW_1(s) \vspace{4pt}\\
\zeta(s=0) & = & \xi_0 .
\end{array}
\right.
\end{equation}
The Fokker-Plank equation associated to this process, which provides a transition probability density $\rho_3$, is
\begin{equation}\label{eq:FP3}
(\partial_s - L_3) \rho_3(\zeta,s|\zeta_0,0) = \delta(\zeta - \zeta_0)\delta(s)
\end{equation}
where the evolution operator is given by $L_3 = X_1 - \kappa^2 X_2^2$.

Each of the three equations (\ref{eq:FPV}), (\ref{eq:FPT}) and (\ref{eq:FP3}) is defined by a Markov generator $\Ll$ of a stochastic process over a manifold $X$, with transition probability $\varrho$ satisfying
\begin{equation}\label{eq:FP}
(\partial_s - \Ll)\varrho(x,s|x_0,0) = \delta(x - x_0)\delta(s),
\end{equation}
and our aim is to use the density $\varrho$ to define a connectivity kernel over the manifold $X$, disregarding the dynamics over the evolution parameter $s$. This connectivity be obtained by integrating $\varrho$ over $s$, against some appropriately chosen weight $p$, hence defining a connectivity kernel as
$$
\Gamma(x,x_0) = \int_0^\infty \varrho(x,s|x_0,0) p(s) ds .
$$
When $p$ is a positive weight normalized to 1, this kernel has the well known probabilistic interpretation of being the transition probability for a stochastic process subordinated to a time process with independent increments distributed with $p$ (see e.g. \cite{Sato99}).
One common choice for the weight $p$ is that of \cite{Mumford, Williams}, that is an exponential decay representing a decay of the signal during the propagation, which amounts to replace $\varrho$ with its Laplace transform at a fixed time-scale. Another possible choice is the one used in \cite{SCS, BCCS} where $p$ was identically set to 1, which amounts to consider the density of points reached at any value of the evolution parameter. In this work we will deal with an intermediate choice between these two, namely we will choose a weight depending on an evolution-scale parameter $H$ as
$$
p(s) = \frac{1}{H} \chi_{[0,H]}(s) = \left\{
\begin{array}{rl}
\frac{1}{H} & s \in [0,H]\\
0 & \textnormal{otherwise} .
\end{array}
\right.
$$
This choice amounts to assign a uniform weight to all values of the evolution parameter, but allows to keep track of the evolution length over which the stochastic paths are evaluated. Our notion of connectivity kernel will then be
\begin{equation}\label{eq:connectivity}
\Gamma^H(x,x_0) = \frac{1}{H}\int_0^H \varrho(x,s|x_0,0) ds . 
\end{equation}
It is worth noting that both the diffusion parameter $\kappa$ and the evolution-scale parameter $H$ could be treated, in principle, as additional fiber variables of the model, linking their different associated connectivities to some prominent feature of the image detected by the visual cortex, for example curvature and scale. For the sake of model simplicity, though, in this paper we will treat these variables as parameters, addressing the extension of the model in a future work.

\subsection{Discrete connectivity kernels}\label{sec:numerics}

In this section we outline the numerical method we have used to compute the connectivity kernels $\Gamma^H$, which are needed due to the lack of analytic solutions. A notable exception is the $SE(2)$ case (\ref{eq:FP3}), for which an analytic solution was obtained in \cite{DvA} and compared in \cite{ZDR} to several numerical approaches, all of them being different from the present one. Our choice of numerics has privileged the robustness and flexibility of the implementation, which has been applied to all the three cases.

The differential equations of type (\ref{eq:FP}) that we need to solve originate from a stochastic process over $X$, being this $\M_0$, $\M_T$ or $\M_3$, that we can write in terms of its sample paths $\gamma : \R^+ \to X$ as
$$
d\gamma(s) = A(\gamma) ds + B(\gamma) dW
$$
where $A$ is a vector field and $B$ is a matrix field over $X$ representing, respectively, (\ref{eq:pathV}), (\ref{eq:pathT}) and (\ref{eq:path3}). A flexible and efficient numerical technique is then that of Markov Chain Monte Carlo methods (MCMC) (see e.g. \cite{MCMC}), which was implemented as follows. We first fix a discretization over the parameter $s$, using without loss of generality a step $\Delta s = 1$ so that the discrete evolution will be performed over $\N$, and a discrete covering grid $\{\Omega_j\}_{j \in \N}$ of $X$, i.e. a collection of subsets of $X$ satisfying $\Omega_i \cap \Omega_j = \emptyset$ if $i \neq j$ and $\bigcup \Omega_j = X$. For a given $\gamma_0 \in X$ we then simulate $N$ several discrete-time random paths over $X$ with the recursive equation
$$
\gamma_{h+1} = \gamma_h + A(\gamma_h) + B(\gamma_h) \delta_h , \quad h \in \N
$$
where $\{\delta_h\}_{h \in \N}$ are (vector valued) i.i.d. gaussian random variables, and assign to each region $\Omega_j$ a value between $0$ and $1$ corresponding to the number of paths that passed through it divided by $N$. This provides a distribution over the cells $\Omega_j$ that, up to a multiplicative constant, for large values of $N$ gives a discrete approximation of the solution to (\ref{eq:FP}) that we will denote $\varrho(\Omega_j,h|\gamma_0,0)$. The resulting discrete approximation of the connectivity kernel (\ref{eq:connectivity}) will then be computed as
\begin{equation}\label{eq:discreteKernel}
\hat{\Gamma}^H(\Omega_j|\gamma_0) = \frac{1}{H}\sum_{h = 0}^H \varrho(\Omega_j,h|\gamma_0,0) .
\end{equation}
A deeper discussion of this kind of numerical approximation for stochastic differential equations can be found in \cite{Platen1999} and references therein.

During the simulations of Section \ref{sec:grouping} we will vary the parameters which characterize the kernels. In order to keep track of their dependency we will then use the notation $\hat\Gamma_0^{H,\kappa,\alpha}$ and $\hat\Gamma_T^{H,\kappa,\alpha}$ for the discrete approximations of the connectivity kernels in the cases, respectively, of (\ref{eq:pathV}) and (\ref{eq:pathT}), where $\kappa$ and $\alpha$ stand for the diffusion coefficients over $\theta$ and $v$, and the notation $\hat\Gamma_3^{H,\kappa}$ for the discrete approximation of the connectivity kernel in the case (\ref{eq:path3}), where $\kappa$ is the diffusion coefficient over $\theta$.
\begin{figure}
\centering
\includegraphics[width=.8\textwidth]{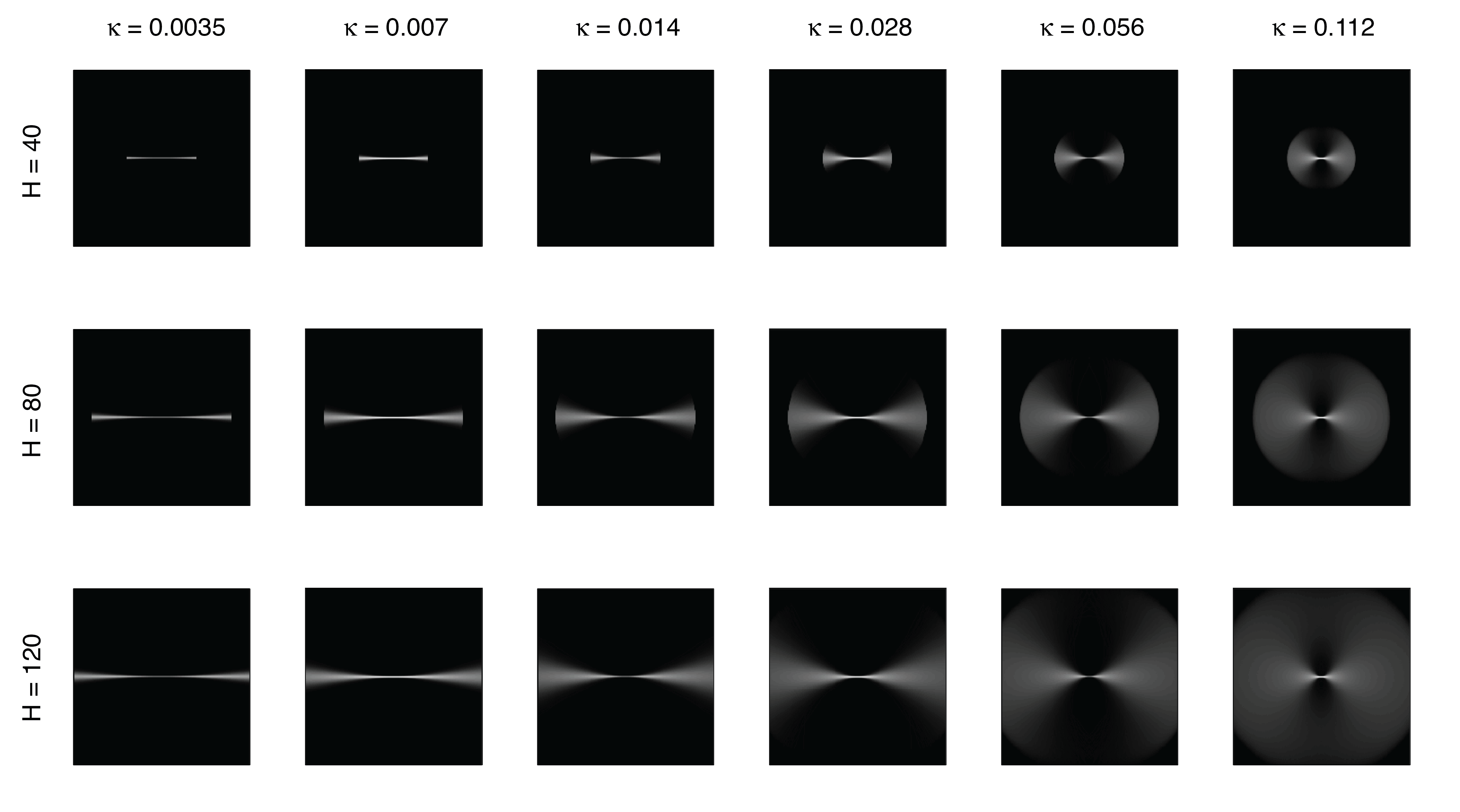}
\caption{Marginal distributions over the $(q_x,q_y)$ plane of the kernel $\hat\Gamma_3^{H,\kappa}$ computed for different values of the parameters $k$ and $H$.}\label{fig:manykernels}
\end{figure}

\begin{figure}
\centering
\includegraphics[width=.48\textwidth]{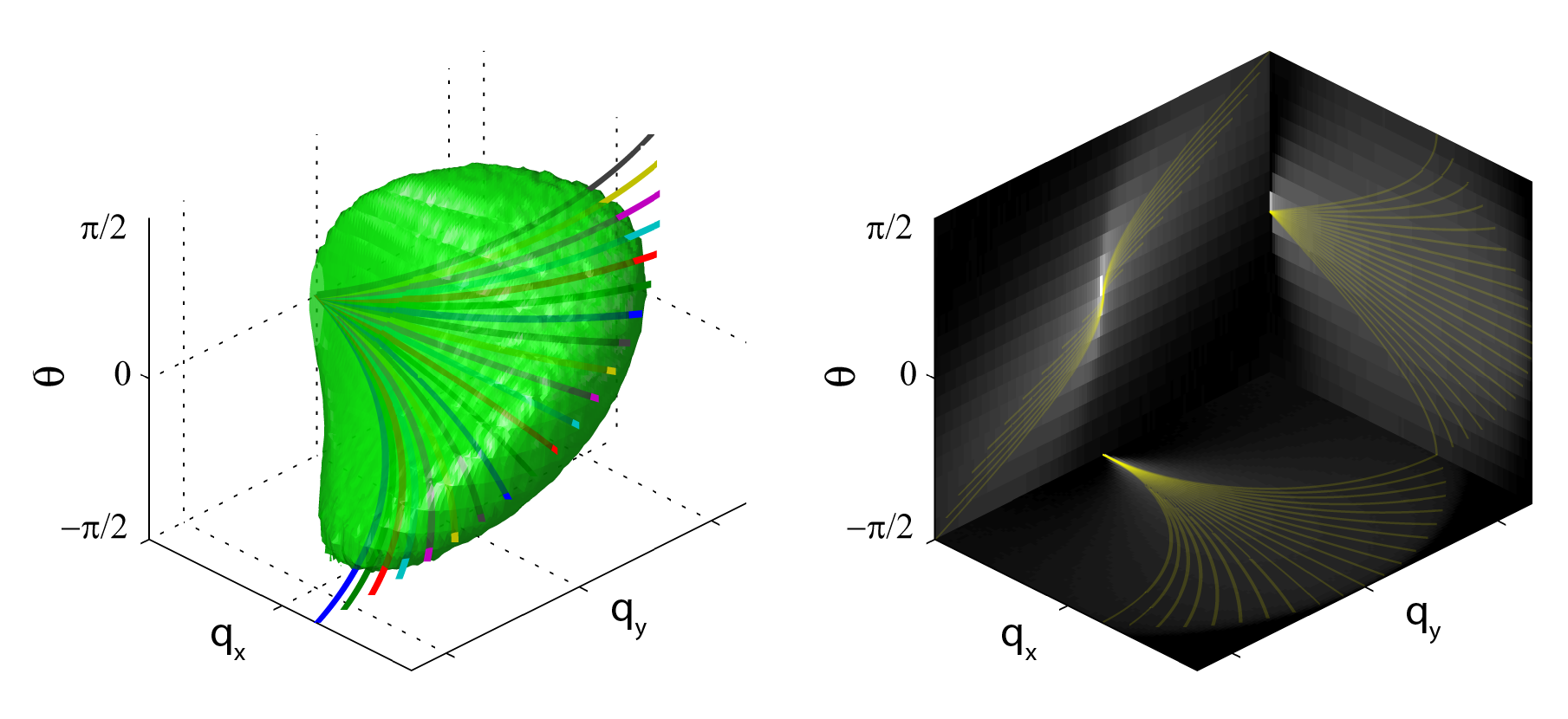} \quad
\includegraphics[width=.48\textwidth]{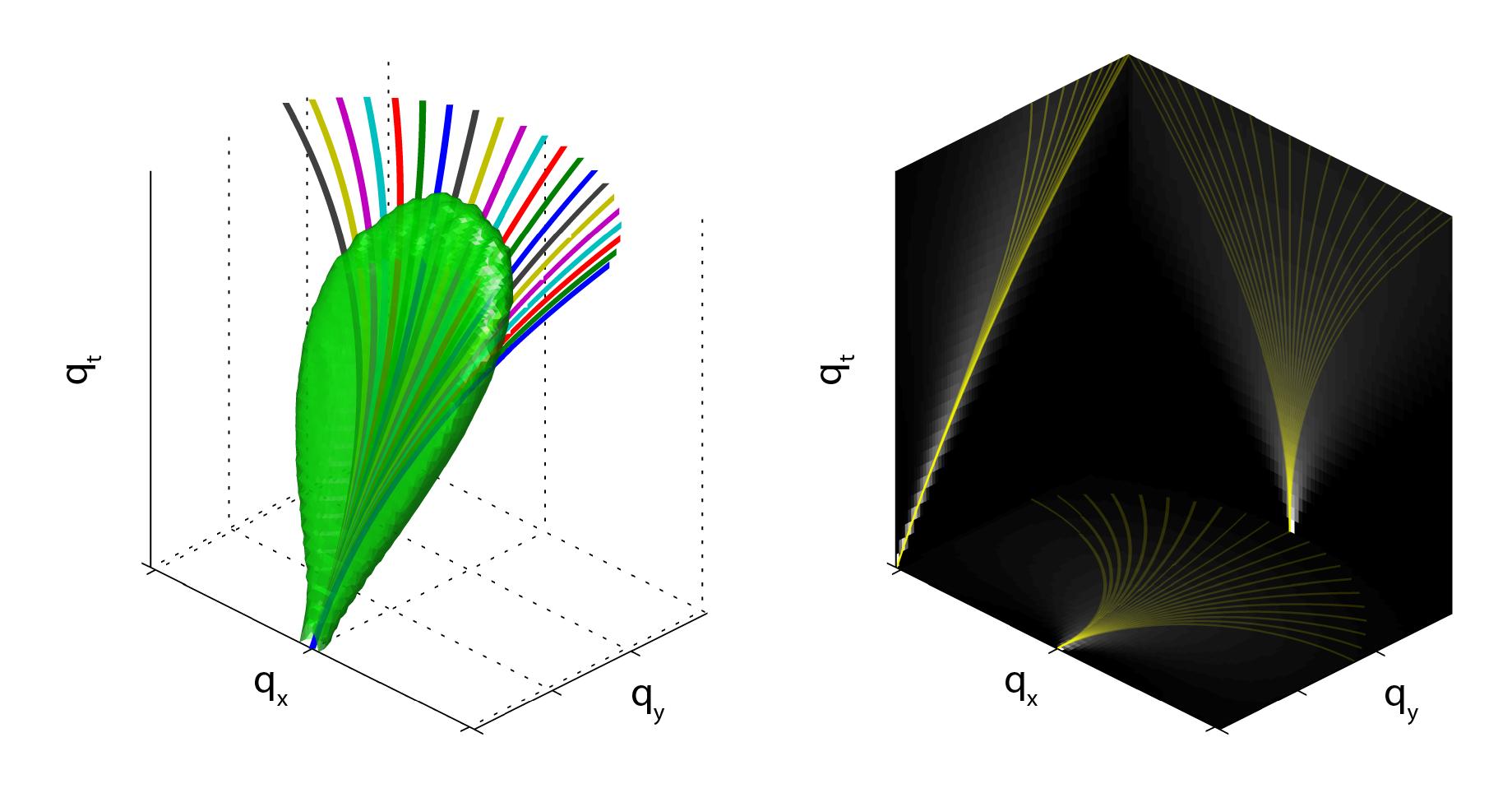}
\caption{Stochastic kernels $\hat\Gamma_0^{H,\kappa,\alpha}$ (top) and $\hat\Gamma_T^{H,\kappa,\alpha}$ (bottom) for $N = 10^6$ stochastic paths, compared to the horizontal curves calculated as in \cite{BCCS}. Left: isosurface plot of the kernels. Right: kernel projections relative to the variables $(q_x,q_y,\theta)$ for $\hat\Gamma_0^{H,\kappa,\alpha}$ and $(q_x, q_y, q_t)$ for $\hat\Gamma_T^{H,\kappa,\alpha}$ (gray), under the projections of the horizontal curves (yellow).}
\label{fig:isokernels}
\end{figure}

The effect of the variation of the parameter $\kappa$ on the shape of $\hat\Gamma_3^{H,\kappa}$ is shown in Fig. \ref{fig:manykernels}, while in Fig. \ref{fig:isokernels} we show the isosurfaces of two projections of the connectivity kernels $\hat\Gamma_0^{H,\kappa,\alpha}$ and $\hat\Gamma_T^{H,\kappa,\alpha}$ together with the related \emph{horizontal curves}, i.e. the curves obtained from the systems (\ref{eq:pathV}) and (\ref{eq:pathT}) by substituting noise with a constant, and varying the parameters $\kappa$ and $\alpha$. These curves were introduced in \cite{BCCS} as the geometric counterpart, in the space-time contact structure, to the horizontal curves used in \cite{CS} to model the orientation association fields. They provide then a natural geometric extension of the notion of association fields to spatio-temporal stimuli with orientation and velocity features. As it can be seen, the related kernels reach their maximum values in the proximity of the fan of such curves originating from the same starting point, which motivates their role as a model for a spatio-temporal neural connectivity.

\section{Spectral analysis of connectivities}
\label{sec:cdm}

In this section we recall the main notions to be used about clustering with spectral analysis on graphs, and how to use the previously introduced geometric setting for these purposes. The task we will address fall into the well-known class of problems broadly referred to as dimensionality reduction, which deals with the general problems of data partitioning and of locality-preserving embeddings of high-dimensional data sets. The literature on these topics is huge, and we have chosen to refer only to works directly related to the present one, addressing the reader to the bibliography contained therein.

\begin{figure}
\centering
\includegraphics[width=.45\textwidth]{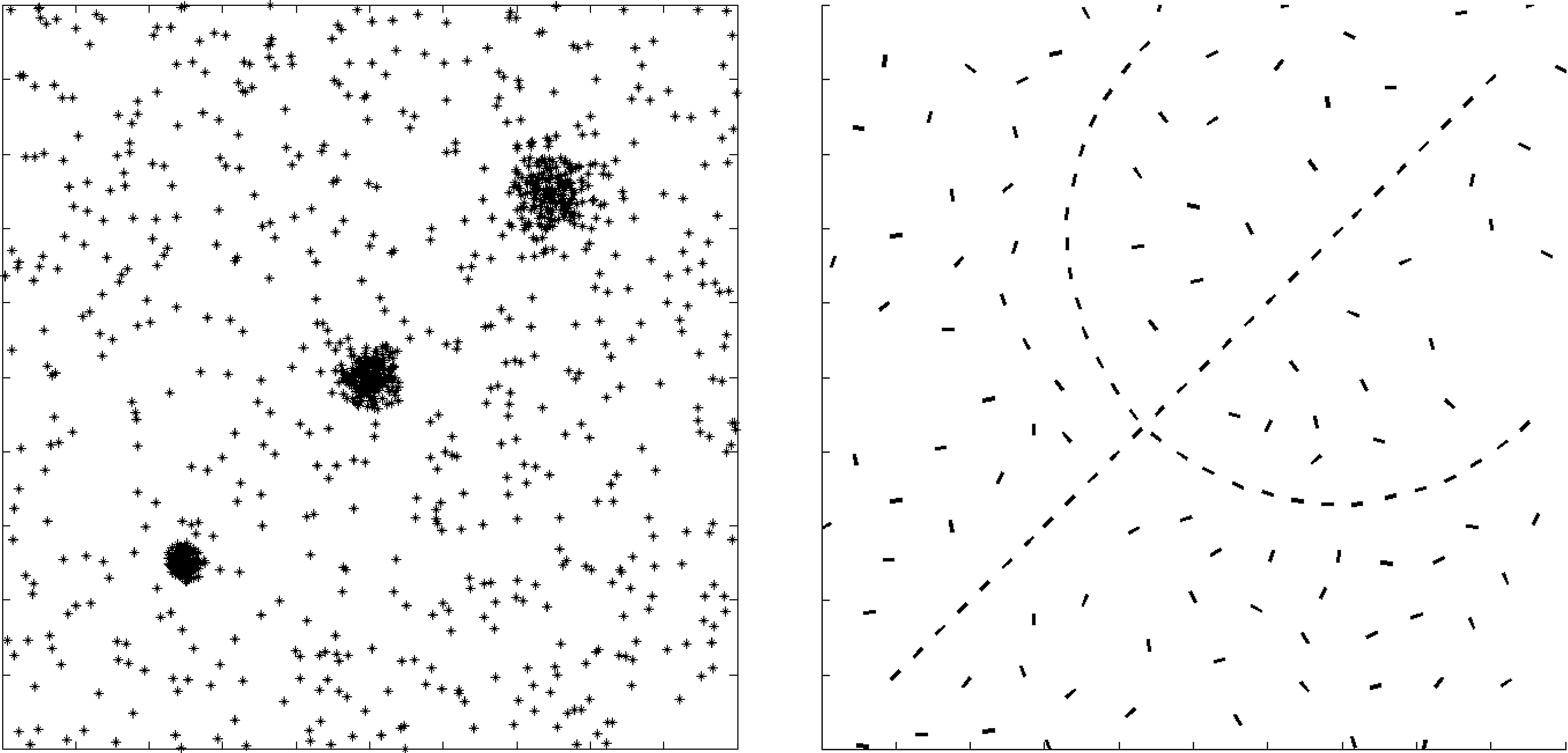}
\caption{Sample spatial stimuli containing perceptual units embedded in random fields.}\label{fig:sampleclus}
\end{figure}

The cognitive task of spatial or spatio-temporal visual grouping can be interpreted as a form of clustering. Two examples of spatial visual stimuli providing standard clustering problems are portrayed in Fig. \ref{fig:sampleclus}. On the left, three dense Gaussian distributions of 2-dimensional points are embedded within a sparser set of random points uniformly scattered throughout the domain. The human visual system normally segregates the points of the Gaussian clouds into three separate groups of points (objects) lying on a noisy environment. On the right, two dashed continuous lines are embedded in a field of segments having random position and orientation. In this case, stimulus collinearity gives rise to a pop-out effect that makes the two lines easily distinguishable from the background, that is the phenomenon quantified by the psychophysical experiments of \cite{Field1993}. 


Although these examples show two quite different grouping effects, a common underlying mechanism can be formalized as follows. Given a data set of $n$ points $S = \{x_i\}_{i = 1}^n \subset X$ living in an arbitrary metric space $(X,d)$, the task of grouping together the points that -- according to the distance $d$ -- are closer, or more \emph{similar} to each other (so that their ensemble forms an \emph{object}), amounts to identify $K$ disjoint subsets $S_i \subset S$, with $i \in \{1,\dots,K\}$, where $K-1$ of them contain points relatively close to each other and relatively far from the rest, and one of them contains points that cannot be classified in such a way, and that will be considered as noise.

It is widely known that this problem is not easily resolvable with purely clustering algorithms like \emph{K-means} (see e.g. \cite{Ng2002}).
A major branch of research has been recently devoted to the development of spectral techniques that allow to address these issues in terms of the spectral properties of symmetric positive semi-definite affinity matrices constructed from the input data set. The ensemble of these techniques can generally be subdivided into two classes \cite{Lafon2006}: methods for locality-preserving embeddings of large data sets, that project the data points onto the eigenspaces of the affinity matrices \cite{Roweis2000,Belkin2003,Coifman2006}, and methods for data segregation and partitioning, that perform an additional clustering step taking as input the projected data set (see e.g. \cite{Perona1998, Weiss1999, ShiMalik, meilashi}).
\newpage
In Subsection \ref{sec:SPC} we will describe the spectral clustering algorithm we will use to perform visual grouping, while in Subsection \ref{sec:CAM} we will show how to use the geometric feature spaces and the cortical connectivities developed in Section \ref{sec:geometry} for the construction of affinity matrices associated to spatial and spatio-temporal visual stimuli. This method will be applied in Section \ref{sec:grouping} to several stimuli.

\subsection{Spectral clustering}\label{sec:SPC}

Let us consider the data set $S = \{x_i\}_{i = 1}^n$ in the metric space $(X,d)$ as the vertices of a weighted graph, where the edge weights $\{a_{ij}\}_{i,1 = 1}^n$ define an \emph{affinity matrix} $A$.
It has been originally shown in \cite{Perona1998} that, when $A$ is a real symmetric matrix, its first eigenvector of can serve as indicator vector for basic grouping purposes. In the same work, the authors also proposed a partitioning algorithm that recursively separate the foreground information from the data set. While this algorithm's implementation is straightforward and efficient, it was demontrated that it can easily lead to clustering errors due to noise, non-linear distributions or outliers \cite{Weiss1999}. This argument was later improved, also in view of its relation to several other problems such as that of minimal graph cuts \cite{ShiMalik}, essentially by performing the spectral analysis of a suitably normalized affinity matrix (see e.g. \cite{tutorial} and references therein). We will use the normalization proposed in \cite{meilashi}, which turns a real symmetric affinity matrix $A$ into the transition matrix $P$ of a reversible Markov chain via row-wise normalization. More precisely, if $D$ is the diagonal degree matrix, having elements
$$
d_{i} = \sum_{j=1}^n a_{i j} ,
$$
the normalized affinity matrix $P$ is given by
\begin{equation}\label{eq:normalizedaff}
P = D^{-1} A .
\end{equation}
This matrix won't be, in general, symmetric, but it can be shown that its eigenvalues $\{\lambda_j\}_{j = 1}^n$ are real and satisfy $0 \leq \lambda_j \leq 1$, and its eigenvectors $\{u_j\}_{j = 1}^n$ can accordingly be chosen with real components.
The clustering properties of the eigenvectors of $P$ can be clearly understood in the following ideal case. Suppose 
that the graph $G = (S,A)$ with nodes given by $S$ and edge weights given by $A$ has $K$ connected components $\{G_i\}_{i = 1}^K$, and that all the elements of each component have the same edge weight connecting them. The resulting normalized affinity matrix $P$ would then be a block diagonal matrix, with only $K$ non-null eigenvalues $\{\lambda_i\}_{i = 1}^K$ each of them equal to $1$, whose corresponding eigenvectors $\{u_i\}_{i = 1}^K$ are piece-wise constant indicator functions of the partitions.

In real applications, the affinity matrices are perturbed versions of the block diagonal ones and do not possess an ideal binary spectra with purely indicator eigenvalues, thus making the partitioning problem generally ill-posed. The best situation one can hope is then that of a good approximation of the ideal case, where the affinity on the dataset is such that there are clusters of points that are strongly connected mainly to their cluster neighbours, and only weakly connected to the rest. In general, several authors demonstrated that the egenvectors $\{u_i\}_{i = 1}^K$ of the normalized affinity matrix $P$, corresponding to the $K$ largest eigenvalues $\lambda_1 \geq \lambda_2 \geq \ldots \geq \lambda_K$, solve the relaxed optimization problem of normalized graph cuts, and give a nice probabilistic interpretation of the clustering problem.

A crucial step in real applications is then that of choosing the value of $K$, that is deciding how many eigenvalues are worth to be taken into consideration and consequently how many eigenvectors possess relevant clustering information. Many authors proposed different solutions, like looking for the maximum eigengap or trying to minimize a particular cost function (see e.g. \cite{Perona2004}). We decided to adopt a semi-supervised solution consisting of fixing an a-priori significance threshold $\epsilon$, and consider as clustering eigenvectors all those $u_i$ whose $\lambda_i > 1-\epsilon$.

\begin{figure}
\centering
\includegraphics[trim=6cm 5cm 4.5cm 3.5cm, clip=true,width=.65\textwidth]{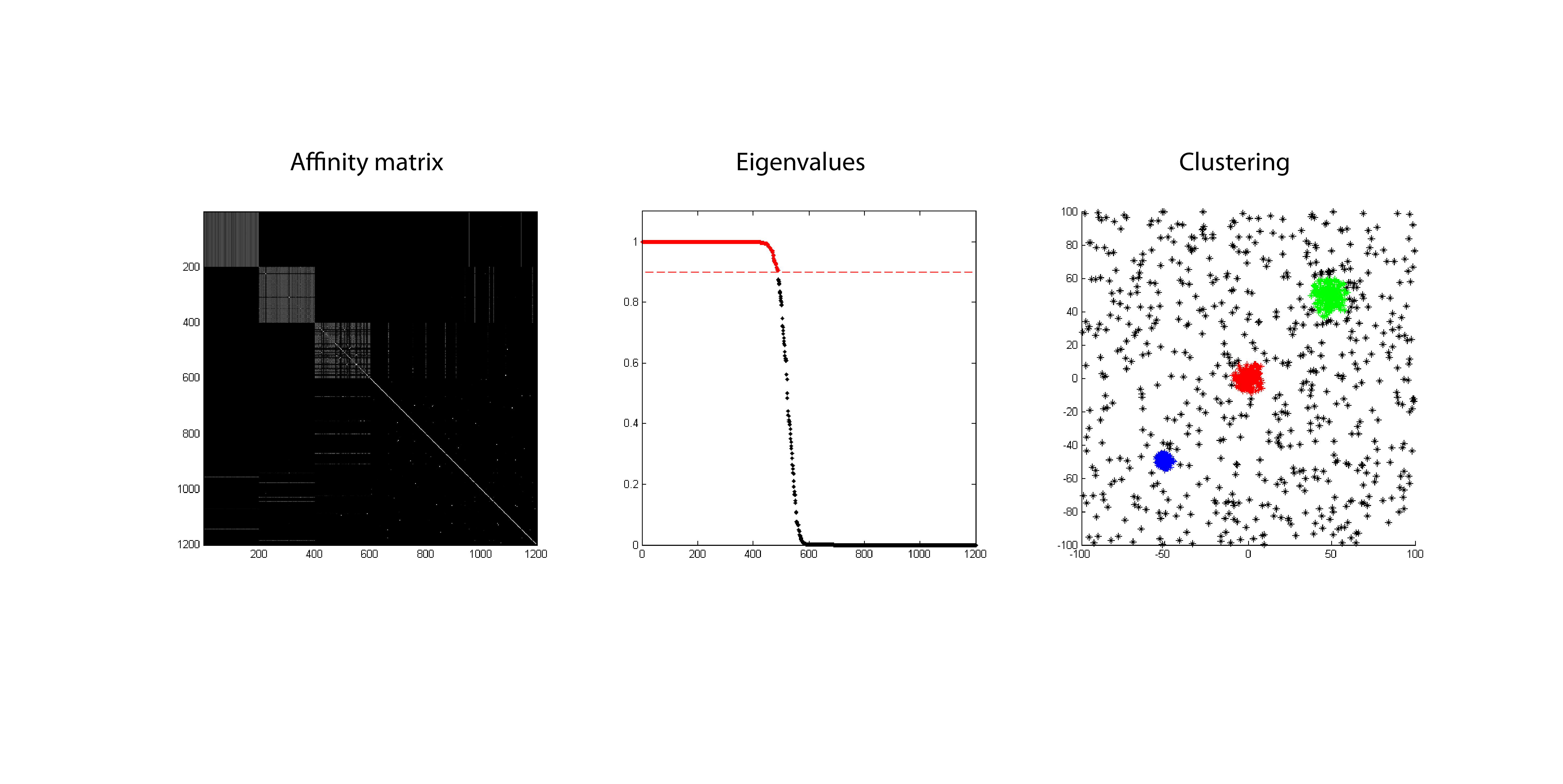}
\includegraphics[trim=6cm 6cm 4.5cm 6.5cm, clip=true,width=.65\textwidth]{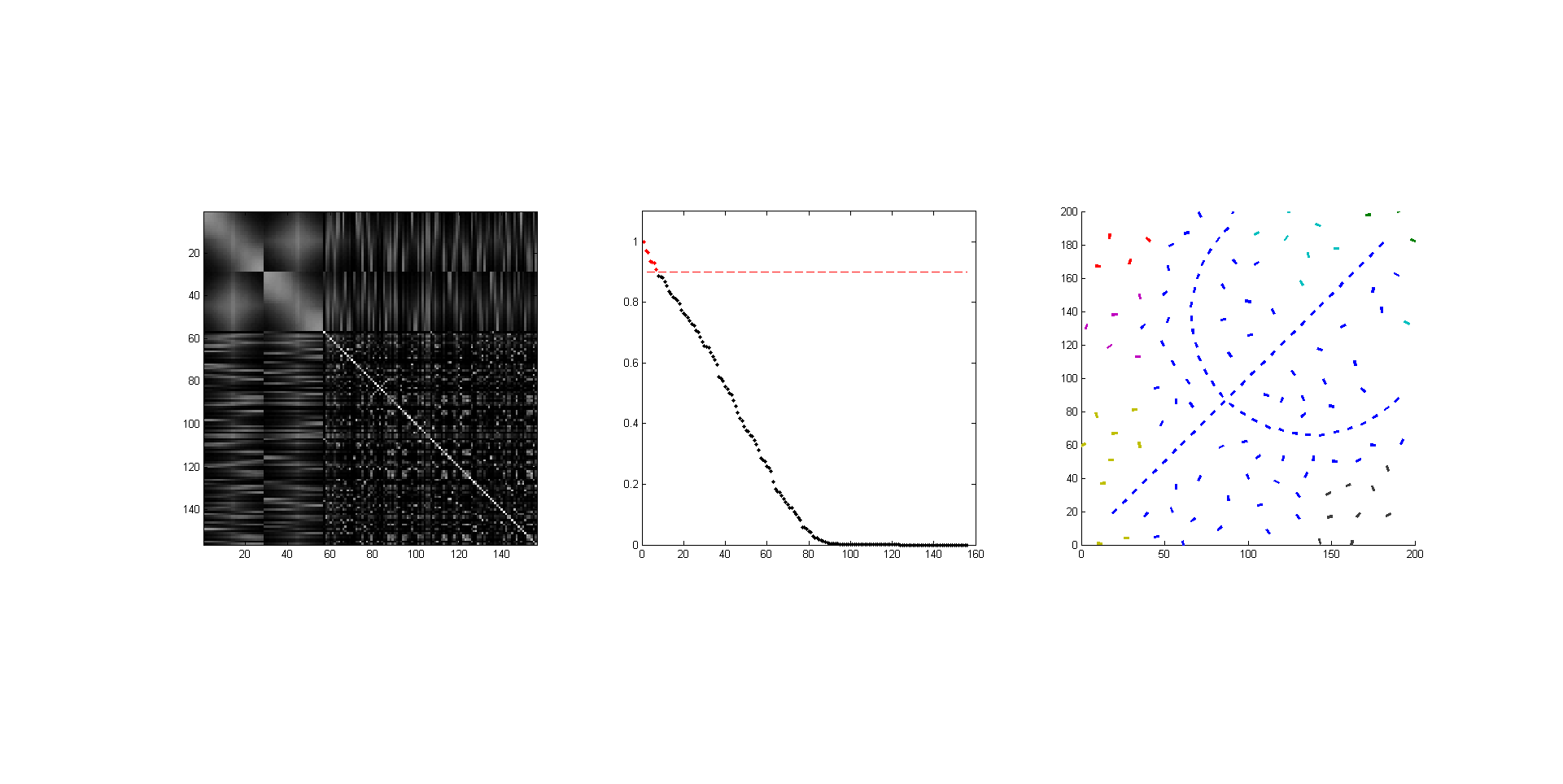}
\caption{Result of the proposed algorithm for the two example data sets. The picture shows how an Euclidean isotropic kernel happens to be an optimal choice to cluster groups of points living in the image plane $\R^2$, but results inadequate when trying to separate boundaries or contours, whose additional feature of local orientation suggests to consider as points on the contact manifold $\M_3 = \R^2 \times S^1$.}\label{fig:sampleclus2}
\end{figure}

Since the more $P$ is far from being similar to a block diagonal matrix, the more its spectrum will be far from being dichotomous, with the ordered $\lambda$'s decreasing more smoothly (see e.g. Fig. \ref{fig:sampleclus2}), in the more ill-posed cases the sensitivity to small changes on the values of $\epsilon$ may become very high. In order to facilitate this delicate passage, we have then used a technique suggested by the well-known diffusion map approach \cite{Coifman2006, Lafon2006}: we have introduced an auxiliary thresholding integer parameter $\tau$, and we have evaluated the exponentiated spectrum $\{\lambda_i^\tau\}_{i = 1}^n$, which for sufficiently large values of $\tau$ is closer to dichotomy, against the threshold $\epsilon$. This exponentiated spectrum has an easy probabilistic interpretation, being the spectrum of the matrix $P^\tau$. Indeed, due to the normalization (\ref{eq:normalizedaff}), $P$ can be seen as the Markov transition matrix of a random walk over the graph, so that $P^\tau$ represents the transition probability of the same random walk in $\tau$ steps.


Once the number $K$ of eigenvectors to use has been selected, we have used a variation of a simple clustering technique proposed in \cite{Kannan} in order to extract the clustering information. It corresponds, in the notation of \cite{Kannan}, to the clustering of the rows of $P$ based on its reduced matrix of eigenvectors $\hat U =  \left[ u_1 \ldots u_q \right]$. The main differences are that, once the thresholding parameters are fixed, this algorithm dynamically assigns the number of (pre)clusters, and that a notion of background is explicitly introduced in the last step. We have summarized our spectral clustering algorithm in Table \ref{tab:algorithm}.



\begin{table}[h!]
\centering
\begin{tabular}{|c|p{.85\textwidth}|}
\hline
1. & \tt Build the affinity matrix $A$ upon an appropriate connectivity measure\\
\hline
2. & \tt Compute the normalized affinity matrix $P = D^{-1} A$\\
\hline
3. & \tt Solve the eigenvalue problem $P u_i = \lambda_i u_i$, where the order of indices $i=1\dots n$ is such that $\{\lambda_i\}_{k = 1}^n$ is decreasing\\
\hline
4. & \tt Fix a threshold $\epsilon$ and a diffusion parameter $\tau$\\
\hline
5. & \tt Define $q = q(P; \epsilon, \tau)$ as the largest integer such that $\lambda_i^\tau > 1 - \epsilon$ for all $i = 1 \dots q$\\
\hline
6. & \tt Define the preclusters $\{\hat C_k\}_{k = 1}^q \subset \R^d$ so that for any $i = 1 \dots n$ the point $x_i$ belongs to $\hat{C}_k$ whenever
$$k = \textnormal{arg}\max_{\!\!\!\!\!\!\!\!\!\! j \in \{1 \dots q\}} \{u_j(i)\}$$
\vspace{-10pt}\\
\hline
7. & \tt Fix a minimum cluster size $M$, join together the preclusters with less than $M$ elements into the cluster $C_0$, and order the remaining preclusters so to obtain a partition of the dataset into $\{C_k\}_{k=1}^K \subset \R^d$, where $K = K(P;\epsilon,\tau,M)$.\\
\hline
\end{tabular}
\caption{Spectral clustering algorithm.}\label{tab:algorithm}
\end{table}
\newpage
As a first application of this algorithm, consider the two data sets presented in Fig. \ref{fig:sampleclus} as sets of points in $\R^2$ endowed with the isotropic Euclidean distance $d(x,y) = \|x - y\|$, and define the affinity matrix
\begin{equation}\label{eq:gaussianAff}
a_{ij} = e^{- \frac{d(x_i,x_j)^2}{2 \sigma^2}} ,
\end{equation}
where $\sigma$ is a scale parameter that has to be chosen upon the characteristics of the data set to be clustered. The result is that of two connected graphs where the similarity between vertices decays as a Gaussian with their Euclidean distance. This is intuitively suitable to describe the visual clustering of the first dataset, but does not take into consideration the kind of similarity that characterizes the visual grouping of the second one. The results of the application of the spectral clustering algorithm with affinity given by (\ref{eq:gaussianAff}) is shown in Fig. \ref{fig:sampleclus2}. In the first case, the algorithm performs the clustering process correctly, finding automatically the number of the main perceptual units and assigning the remaining elements to the noise/background cluster. It is worth noting that the spectrum of the normalized affinity matrix $P^\tau$ counts many eigenvalues that are close to $1$, each only representing a single perceptual unit, that are mostly composed of few elements. The segment data set, on the other hand, was not clustered correctly: this was predictable, as these kind of stimuli are characterized by a local feature of orientation and hence should be better considered on the position-orientation domain $\R^2 \times S^1$, together with an anisotropic affinity.


\subsubsection{On possible neural implementations of the algorithm}

Besides the purely computational presentation that we have given of our spectral clustering algorithm, it may actually be relevant to consider whether plausible neural computations exist that could be responsible of its implementation in the visual cortex.

The discussion of Section 2 and the work \cite{BCCS} on the psychophysical and physiolological nature of the connectivity kernels constitute our main motivation to consider the first step of the algorithm as a cortical implementation of affinities in the feature spaces. On the other hand, the normalization step 2 can be understood as formally analogous to the one introduced in \cite{Tononi94}, and may be motivated by the evidence that biological neural systems tend to adjust the weight of afferent connections so that a neuron with few incoming connections will weight those inputs more heavily than a neuron with many incoming connections (see also the discussion in \cite{Barnett09}).

With respect to step 3, previous works such as \cite{Bressloff02, Faugeras09} describe a possible implementation of the spectral analysis as a mean-field neural computation, obtaining the emergence of eigenvectors of the connectivity as a symmetry breaking in the evolution equation associated to sufficiently high eigenvalues. An extension of these works can be found in \cite{SartiCittiPercepts}, where it is shown that in the presence of a visual input the emerging eigenvectors correspond to visual perceptual units, which are thus obtained from a spectral clustering on excited connectivity kernels. According to these models, when the magnitude of an eigenvalue trespasses a given threshold, its corresponding eigenvector becomes a locally unstable solution, hence generating a distinguishable activity pattern. This threshold is what we aim to reproduce with steps 4 and 5.

Coming to step 6, we first note that its combinatorial formulation represents a concrete implementation of the following principle: eigenvectors with high eigenvalues represent preceptual units, and a point of the visual stimulus is assigned the unit whose eigenvector component over such a point has the higher magnitude. As we will discuss more in detail below, such a clustering will perform better the more the affinity matrix is close to a block diagonal matrix, since in such cases the eigenvectors relative to high eigenvalues will be close to indicator functions of weakly overlapping regions of the stimulus. In these situations the concrete neural implementation of steps 3, 4 and 5 already includes our step 6. Indeed, the computation of an almost-indicator eigenvector that is an unstable solutions to the neural field equation means that the underlying cells have been selected and participate to the associated activity pattern. We also observe at this point that in this normalized model all sufficiently high eigenvalues have very close magnitudes, that in the ideal case is 1 for all of them. If a neural field model with two populations is considered, the magnitude of the eigenvalue is associated to the frequency of oscillation of the neural population, which in this situations is almost the same for all perceptual unit and lies in the gamma range. This aspect marks a substantial difference with other non-normalized models of spectral neural computations.

The aim of step 7 is to measure  the \emph{saliency} of the perceptual units, and to introduce a thresholding on it. The notion of saliency that we consider is proportional to the total neural activity involved in the detection of a perceptual unit that, in the special artificial cases under consideration, corresponds to the number of related elements of the dataset.

\subsection{Cortical affinity matrices}\label{sec:CAM}

As we have described in Section \ref{sec:geometry}, a spatial or a spatio-temporal visual stimulus is represented by different classes of V1 specialized neurons on different kinds of higher dimensional feature spaces, and we have focused on three of them that we have called $\M_0$, $\M_T$ and $\M_3$. The neural response (\ref{eq:filtering}) is in general non null on the whole feature space, but according to the stronger or weaker presence of the locally preferred feature, the magnitude of the response will be higher or lower. Here we will adopt a great simplification, and consider the response magnitude to be allowed to take only value $0$ or $1$, so that the datasets we will deal with will be purely point datasets embedded in a feature space. The main advantage of this simplification is the possibility to test the grouping capabilities of the geometric connectivities in the feature spaces without dealing with the many delicate questions related to the stimulus representation itself. On the other hand, only a few classes of stimuli are feasible to be represented in such a way. For this reason we will work only with synthetic stimuli and generate the corresponding datasets directly in the feature space. This indeed seems to be the safest way to separate the problems related to the geometric properties of the proposed connectivity from the ones related to the good representation of stimuli onto the feature space.

Let then $X$ be a feature space, such as $\M_0$, $\M_T$ or $\M_3$, let $\{\Omega_j\}_{j \in \N}$ be a discrete covering such as the one introduced in Subsection \ref{sec:numerics}, and let $S = \{x_i\}_{i = 1}^n \subset X$ be the dataset representing the visual stimulus on this higher dimensional space, being this of type $S_0 = \{\xi_i\}_{i = 1}^n \subset \M_0$, $S_T = \{\xi_i\}_{i = 1}^n \subset \M_T$ or $S_3 = \{\zeta_i\}_{i = 1}^n \subset \M_3$. Our aim is to define an affinity that makes two points $x_i$ and $x_j$ more similar the higher is their geometric connectivity, computed as in (\ref{eq:connectivity}) with $\hat\Gamma^H(x_i,x_j)$. A necessary compatibility condition between the dataset and the discretization of the space is that each $\Omega_j$ contains at most one point $x_i$ of the dataset. This indeed makes the kernels well defined kernels on the dataset, and allows us write $\hat\Gamma^H(x_i,x_j)$. However, since the associated Markov generator $\Ll$ is not selfadjoint, this kernel is not symmetric. We will then need to construct symmetric affinities from these connectivities, or rather adapt the previously described theoretical setting to couple with this asymmetry: we will choose one or the other way depending on the kernel, depending on the geometry of the information carried.

\subsubsection{Reciprocal connectivities for spatially distributed features}\label{sec:symmetric}
A reasonable neural assumption when modeling long-range horizontal connections among V1 cells on $\M_3$ or on $\M_0$, which refer to the spatially distributed features of local orientations and local velocities, is that these connections are reciprocal \cite{Bosking1997, Kisvarday1997}. This means that two cells that are selective for the features $(\theta_1,v_1)$ and $(\theta_2,v_2)$ are assumed to be symmetrically connected. We will then take as affinities $A^H = (a^H_{ij})$ a symmetrization of the corresponding connectivity kernels. The symmetrization we have chosen consists of taking the Hermitian component of the kernels:
\begin{equation}\label{eq:corticalaff}
a^H_{ij} = \frac{\hat\Gamma^H(x_i,x_j) + \hat\Gamma^H(x_j,x_i)}{2} .
\end{equation}
This symmetrization has indeed a specific geometric meaning, which can be easily understood for the kernel $\hat\Gamma_3^{H,\kappa}$: it is equivalent to sum the fundamental solution of (\ref{eq:FP3}) with the fundamental solution of the same operator under an angular shift of 180 degrees in the variable $\theta$. Such rotation turns the drift term $X_1$ into $-X_1$, hence transforming the forward Kolmogorov equation into the corresponding backward equation, so the sum of the two solutions is clearly symmetric. On the other hand, that sum allows to identify angles up to 180 degrees, hence turning a process which was a priori defined over the group $SE(2)$ of positions and angles into a process properly defined on positions and orientations. For this reason, such symmetrization is customary when dealing with such nuclei with the purpose of describing pure orientation (see e.g. \cite{SCS}). We observe that this symmetrization does not modify the single cell response of modeled simple cells, which still detect angles, but rather introduces a symmetry in their geometric connectivities so that a cell having a preferred angle is considered to be long range connected to other cells along the two angles corresponding to the associated orientation. The same argument applies to the kernel $\hat\Gamma_0^{H,\kappa,\alpha}$, whose Fokker-Planck operator differs only for an additional diffusion term on the velocities.

Depending on the application to clustering with the kernels $\hat\Gamma_0^{H,\kappa,\alpha}$ or $\hat\Gamma_3^{H,\kappa}$ we will then obtain from (\ref{eq:corticalaff}) different affinity matrices over datasets $\{\xi_j\} \subset \M_0$ or $\{\zeta_i\} \subset \M_3$. The associated spectral clustering will then depend on the parameters $(\epsilon, \tau)$ and $M$ from the algorithm, and on the parametes $H$, $\kappa$ and $\alpha$ from the kernels.

\begin{figure}
\centering
\includegraphics[trim=6cm 5cm 4.5cm 3.5cm, clip=true,width=.95\textwidth]{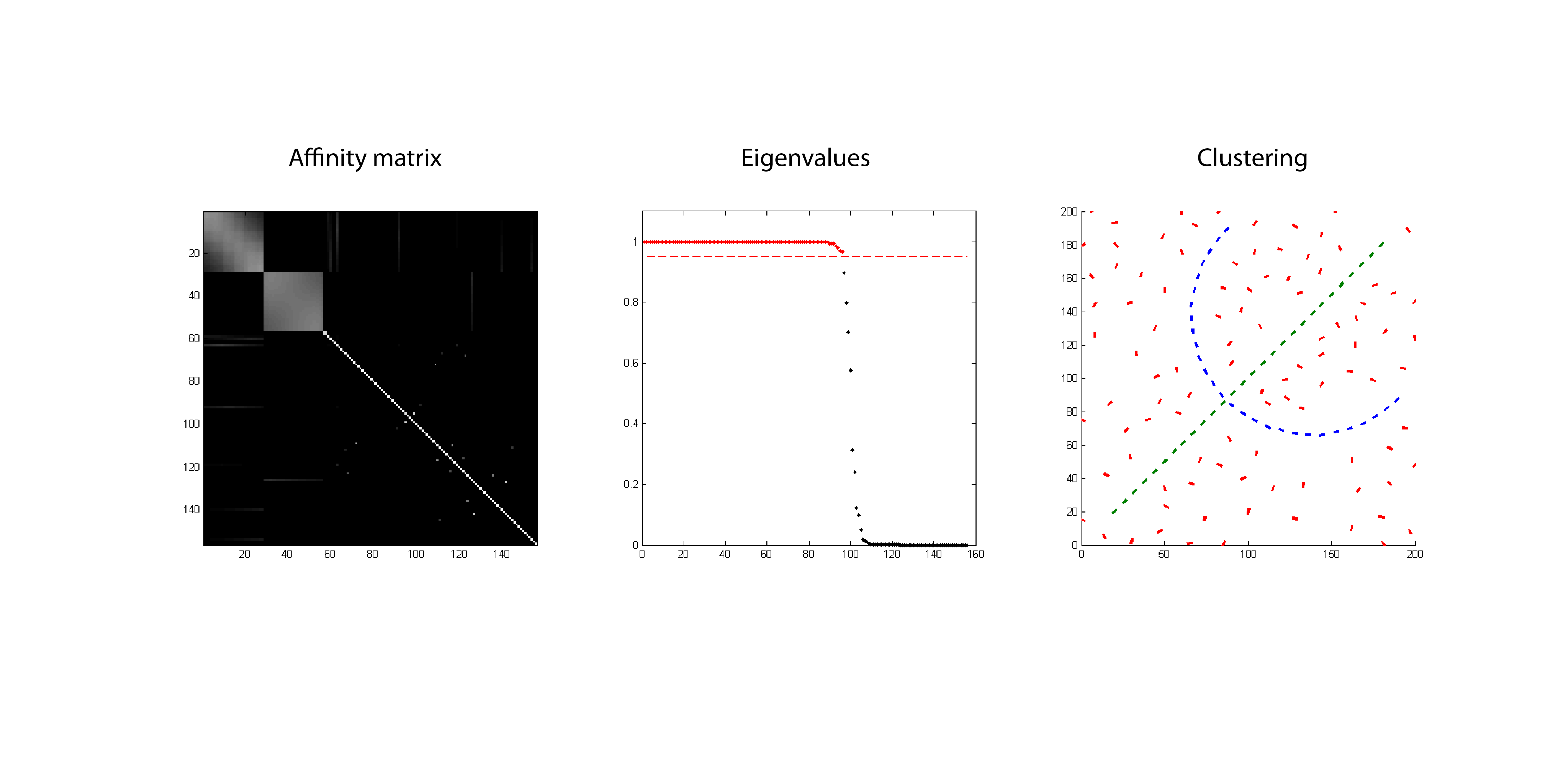}
\includegraphics[trim=6cm 6cm 4.5cm 6.5cm, clip=true,width=.95\textwidth]{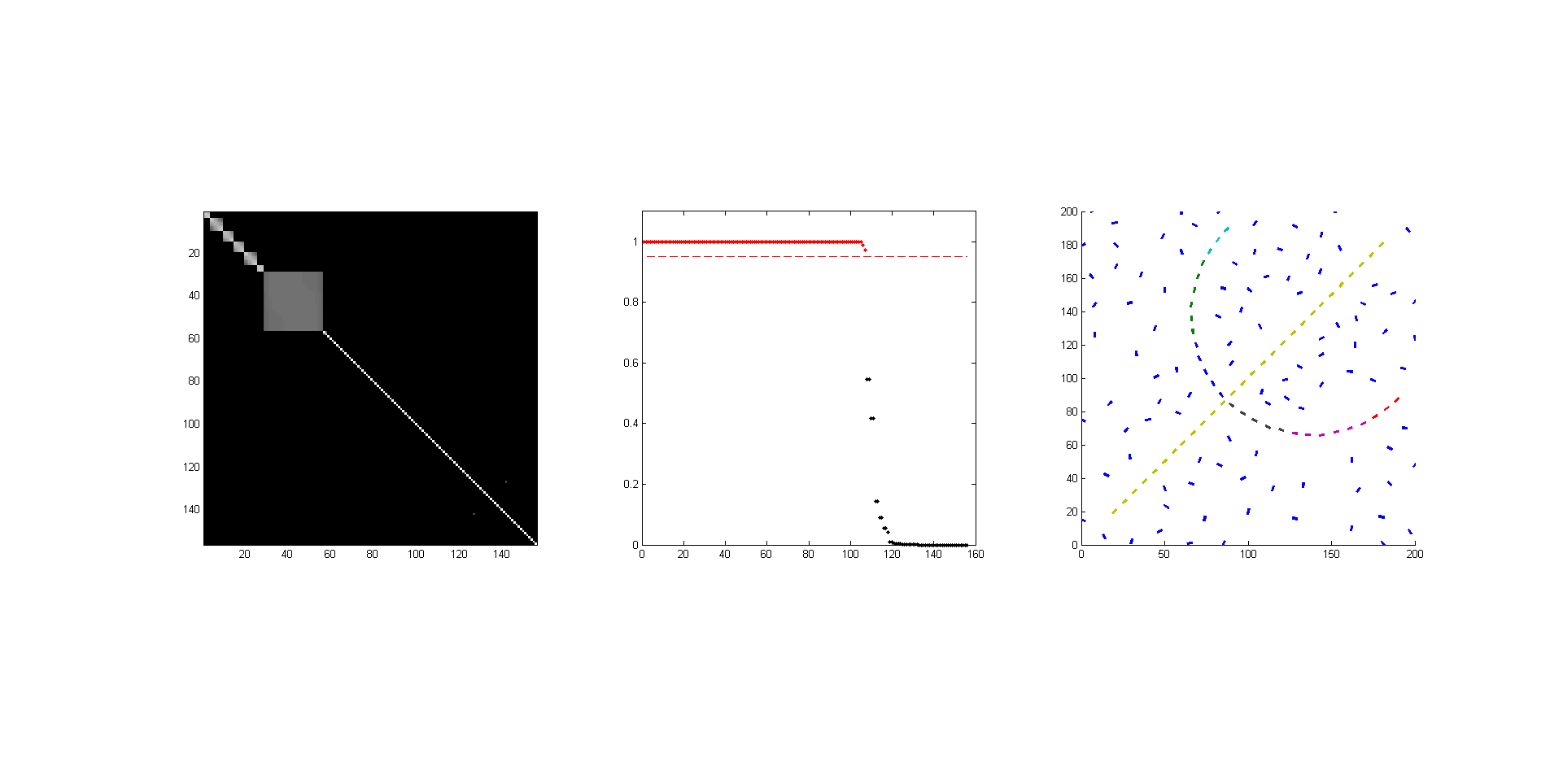}
\includegraphics[trim=6cm 6cm 4.5cm 6.5cm, clip=true,width=.95\textwidth]{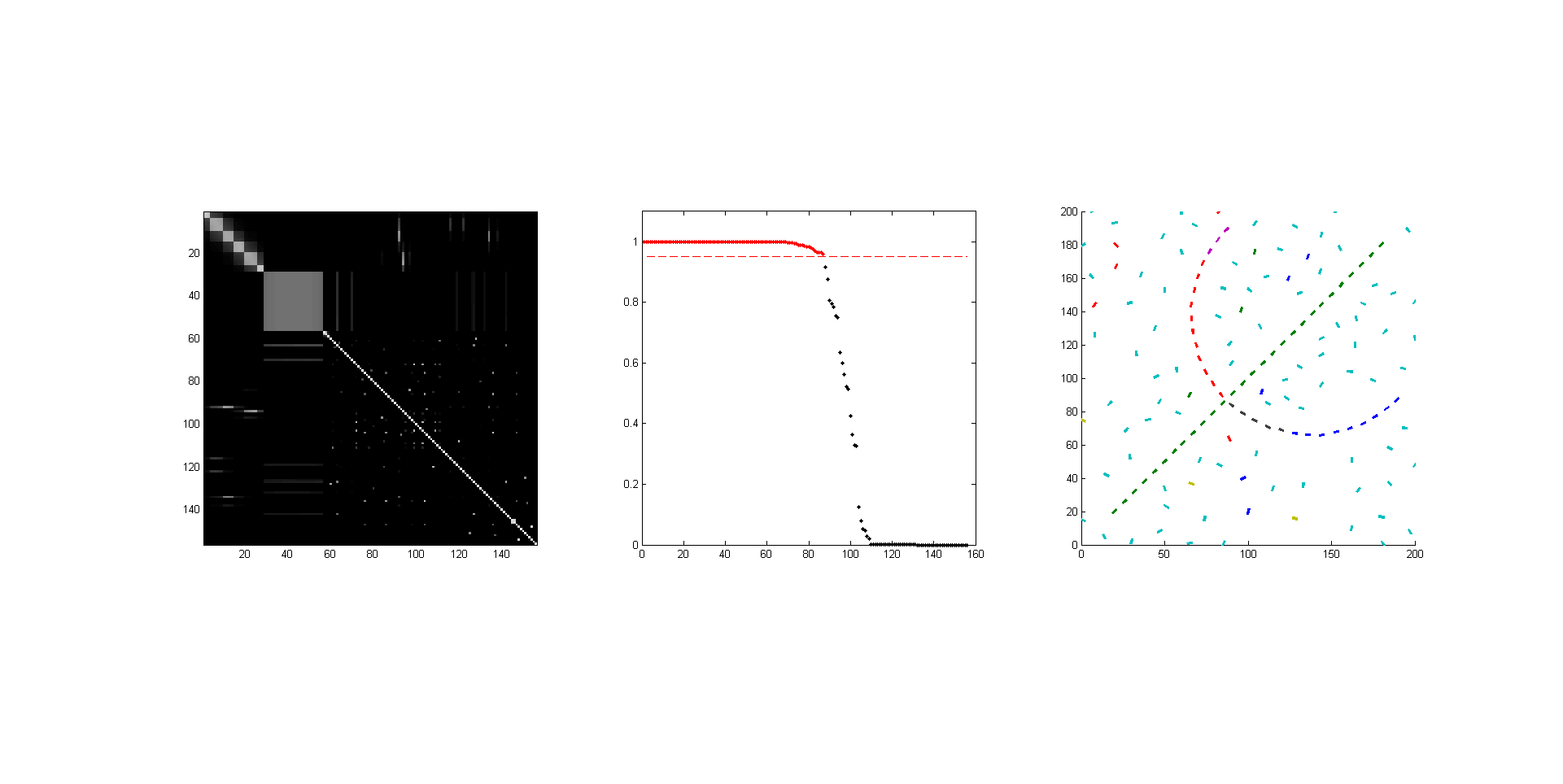}
\caption{Result of the proposed algorithm for the second example data set and different parameters for the kernel $\hat\Gamma_3^{H,\kappa}$. Different kernel parameters modify the look of the affinity matrix, of its spectrum, and of the resulting data set partitioning.}\label{fig:sampleclus3}
\end{figure}

The grouping capabilities of such anisotropic affinities, and the role of the parameters, may be first evaluated on the simplest connectivity $\hat\Gamma_3^{H,\kappa}$. 
In order to do so, we have applied the spectral clustering algorithm with the cortical affinity to the second stimulus of Fig. \ref{fig:sampleclus}. Such stimulus was indeed considered as a dataset in $\M_3 = \R^2 \times S^1$, each segment having a position in $\R^2$ determined by its center, and a position in $S^1$ corresponding to its orientation. The results obtained are displayed in Fig. \ref{fig:sampleclus3} for different sets of kernel parameters $H$ and $\kappa$. The clustering parameters used here and in the examples of the next sections were $\epsilon = 0.05$, $\tau = 150$ and $M = 3$: we have kept those parameters fixed for all the tests, in order to focus on the differences in the grouping properties due to the connectivity kernels.

As we can see, the presence of an affinity that is more semantically and geometrically adapted to the dataset with respect to (\ref{eq:gaussianAff}) positively influences the grouping capabilities of the method. However, already this first simulation shows the relevance of the kernel parameters on the quality of the grouping.

For the top plots of Fig. \ref{fig:sampleclus3}, we have used an evolution integration step of $H = 40$ and an orientation diffusion coefficient $\kappa = 0.014$, which coincides with the curvature of the semi-circular object. The algorithm clearly succeeds in distinguishing the two perceptual units from each other, and correctly assigns the remaining elements to the same background/noise partition.
For the middle plots, we have reduced the value of the diffusion constant to $\kappa = 0.0035$: while the algorithm correctly retrieves the straight contour and distinguish the units from the background, the semi-circle gets over-partitioned. As in the previous case, the affinity matrix is close to being a block diagonal matrix, but in this case the partitioned connected components of the sub-graph containing the two objects are more than two. Setting again the diffusion coefficient to the original value, for the bottom plot we have increased the evolution integration step value to $H = 100$. As the stimulus domain is $200\times 200$ pixel wide, this means that every segment in the example could potentially have a non-null connectivity value with almost an half of the other segments, if the co-circularity conditions are satisfied. Indeed, by observing the resulting affinity matrix $A_3^{H=100,\kappa=0.014}$, we can see high affinity values between the objects and the random noise and in between the background elements. Again, even if the straight line is correctly retrieved as a single object, two random elements, approximately co-circular with the beginning of the line, are uncorrectly interpeted as being part of the object. Furthermore, the semi-circular contour gets again over-segmented, and many of the randomly collinear points, very far from each other, are interpreted as being a perceptual unit.

\begin{figure}
\centering
\includegraphics[width=.7\textwidth]{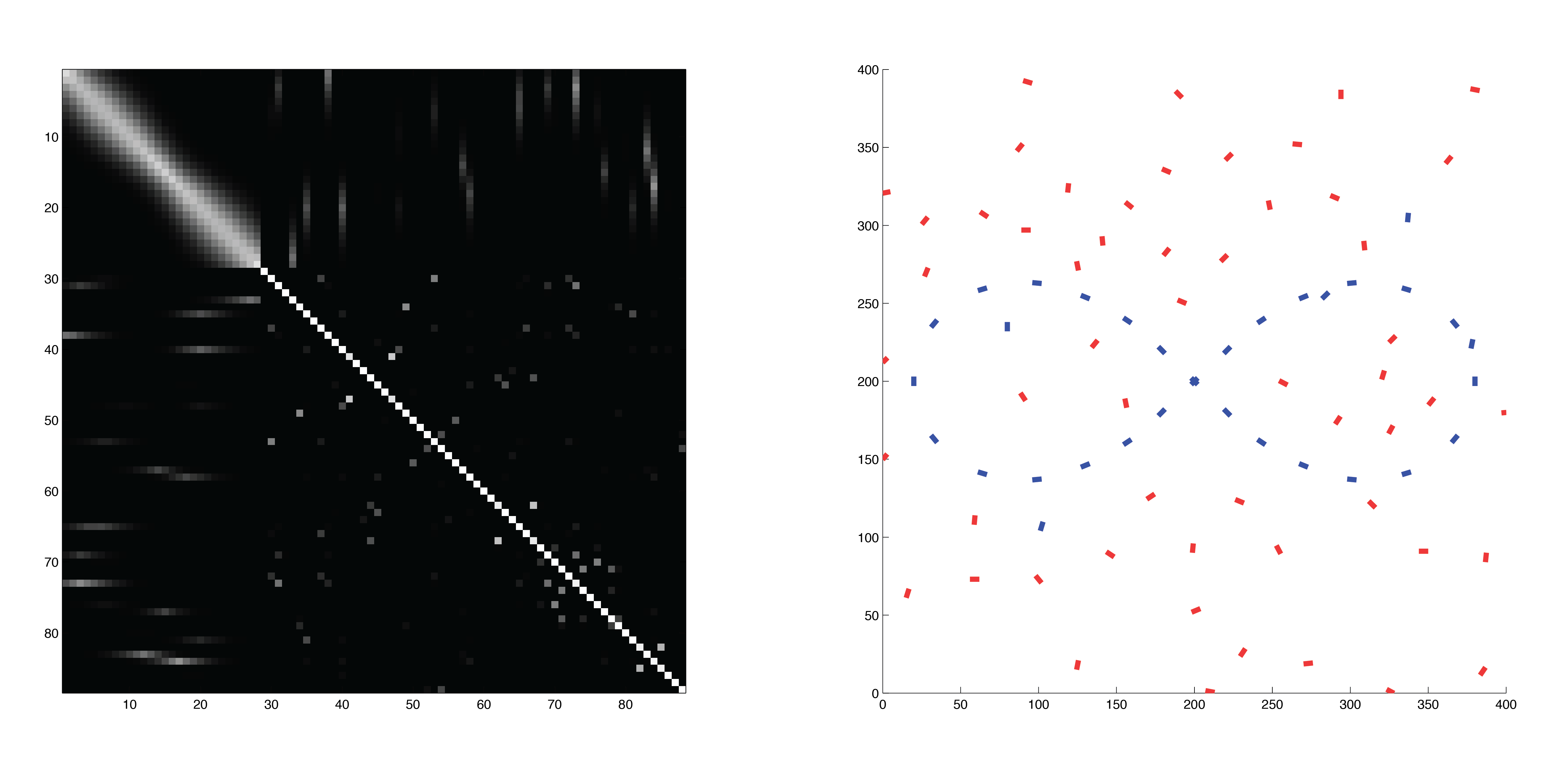}
\caption{Affinity matrix and grouping results of a lemniscate with random noise.}\label{fig:lemniscate}
\end{figure}

In any case, we would like to stress that these kernels do not prevent contours having multiple orientations in the same spatial position to be recognized as one object. See for example Fig. \ref{fig:lemniscate}, with a lemniscate embedded in a field of randomly oriented segments. On the lifted space where the kernel lives, this 8-figure contour is indeed continuous and non-overlapping, so that the grouping algorithm will assign all of its elements to the same perceptual unit, as human vision would tend to do.

\subsubsection{Time asymmetry: a directed graphs approach}\label{sec:asymmetric}

In order to set an affinity over a space-time dataset $S_T = \{\eta_i\}_{i = 1}^n \subset \M_T$ one must consider the intrinsic notion of causality associated to the time component of $\eta$, which from the neural point of view represents the mean activation time of a V1 cell. This causality is reflected in the kernel $\hat\Gamma_T^{H,\kappa,\alpha}$, so that any symmetrization would compromise the information content carried by such kernels and negatively affect the efficiency of the grouping algorithm. We will then need to work with a \emph{nonsymmetric affinity matrix} $A_T^{H,\kappa,\alpha}$, hence defining a \emph{directed graph}, given by
\begin{equation}\label{eq:corticalaffasymm}
(a_T^{H,\kappa,\alpha})_{ij} = \hat\Gamma_T^{H,\kappa,\alpha}(\eta_i,\eta_j) .
\end{equation}
In order to cope with this asymmetry, we will modify some of the spectral clustering criteria. We will still first normalize the affinity matrix as in (\ref{eq:normalizedaff}). This will produce a transition probability matrix $P_T$, which however is associated to a random walk that does not satisfy the detailed balance condition, because the Markov chain is not anymore reversible. Its eigenvalues $\{\lambda_i\}_{i = 1}^n$ and eigenvectors $\{u_i\}_{i = 1}^n$ will then be, in general, complex valued. In order to perform spectral clustering in this situation we will follow the approach introduced in \cite{Pentney2005}, which consists concretely in replacing the eigenvectors $\{u_i\}$ with vectors obtained by the sum of their real and imaginary parts $\{u_i^+ = \Re{u_i} + \Im{u_i}\}$, and defining their clustering strength in terms of the square modulus of their eigenvalues, hence performing the thresholding step $5.$ of the algorithm in Table \ref{tab:algorithm} with respect to the set of real numbers $\{|\lambda_i|^2\}_{i = 1}^n$.

This is by far not the only possible choice. In particular this choice will not preserve the interpretation of minimal graph cuts, but rather it will produce clusters with the property of having elements with approximately the same outward transition probability. Indeed, as discussed in \cite{Meila2007}, the minimal cuts clustering and the probabilistic clustering have in general different solutions for directed graphs, which sets a marked difference with the undirected case associated to symmetric affinities.

The choice of working with the couples $\{(u_i^+,|\lambda_i|^2)\}$, that are not in general eigenva\-lue-eigenvector pairs of a Hermitian matrix, can be better understood as follows. We can see that the eigenvalue problem associated to the real $n \times n$ matrix $P = D^{-1}A$ for a nonsymmetric affinity $A$ can be restated equivalently as an eigenvalue problem with real eigenvectors, by doubling the dimension of the space and replacing $P$ by $\displaystyle \tilde P = \matr{P}{0}{0}{P}$. In a more formal way, let
\begin{equation}\label{eq:eigen}
P u_\lambda = \lambda u_\lambda \ , \quad P \overline{u_\lambda} = \ol{\lambda}\,\ol{u_\lambda}
\end{equation}
and call $\lambda = x + i y$. Let us introduce the following notation: set $\displaystyle\vec{u}_\lambda = \binom{u_\lambda}{\ol{u_\lambda}} \in \C^{2n}$, denote with $\displaystyle \tilde P = \matr{P}{0}{0}{P} = \Id{2} \otimes P$, and let also $\Lambda_\lambda = \matr{\lambda}{0}{0}{\ol{\lambda}} \otimes \Id{n}$, where $\otimes$ is the usual Kronecker product and $\Id{n}$ stands for the $n \times n$ identity matrix. Then solving the problem (\ref{eq:eigen}) for $u_\lambda \in \C^n$ is equivalent to solving the problem
\begin{equation}\label{eq:eigenv}
\tilde P \vec{u}_\lambda = \Lambda_\lambda \vec{u}_\lambda 
\end{equation}
for $\vec{u}_\lambda \in \{(z_1,z_2) \in \C^{2n} : z_2 = \ol{z_1}\}$.

Consider now the auxiliary real vectors $\displaystyle\vec{v}_\lambda = \binom{u^+_\lambda}{u^-_\lambda} = \binom{\Re{u_\lambda} + \Im{u_\lambda}}{\Re{u_\lambda} - \Im{u_\lambda}}$. If we introduce the matrix $B = \matr{\frac{1 - i}{2}}{\frac{1 + i}{2}}{\frac{1 + i}{2}}{\frac{1 - i}{2}} \otimes \Id{n}$, we have that $\displaystyle\vec{v}_\lambda = B \vec{u}_\lambda$, and we can rewrite (\ref{eq:eigenv}) as the \emph{real system} for $\vec{v}_\lambda \in \R^{2n}$
\begin{equation}\label{eq:eigenr}
\tilde P \vec{v}_\lambda = \Sigma_\lambda \vec{v}_\lambda 
\end{equation}
where $\displaystyle\Sigma_\lambda = \matr{x}{y}{-y}{x} \otimes \Id{n} = B \Lambda_\lambda B^{-1} \in \R^{2n \times 2n}$. The magnitude of the action of $\tilde P$ over $\vec{v}_\lambda$ is given by $\det \Sigma_\lambda = |\lambda|^2$, which provides then a natural clustering parameter to threshold. Moreover, since the matrix $\tilde P$ is a double copy of the matrix $P$, in order to cluster its rows it is sufficient to work on a one-half dimensional space and choose, without loss of informations, e.g. only the $u^+_\lambda$ component of $\vec{v}_\lambda$.

\section{Visual grouping with cortical affinities}\label{sec:grouping}

Several phenomenological findings indicate that the grouping properties obtained by spatial collinearity can easily be broken if one associates a speed and an orthogonal direction of movement to each oriented segment. Limits have been found on the maximum rate of change of local speed along a contour that makes possible the perception of boundaries and shapes. Indeed, a random speed distribution over a dashed line could completely destroy the perception of a single unit as a whole, while enhancing the impression of different segments pertaining to the random background field.

These observations have led to the notion of \emph{motion contour} in \cite{Rainville2005}, where it is shown by psychophysical experiments how the brain groups features together also relying on the local speed perpendicular to their orientation axis, with coherent velocities being represented by velocity fields that vary smoothly over space. The former study expanded the already known notions that local stimulus velocity is discernible (thus determinant for grouping purposes) only when it is orthogonal to the perceived contour or it is not part of a trajectory \cite{Hess2003, Ledgeway2005, Verghese2006}. Coherently, the analysis carried out in \cite{CBS} over a data set of cortical neurons in the primary visual cortex showed how the spatio-temporal shape of their RPs is biased to optimally measure the local stimulus orthogonal velocity. Thus, it may be inferred that stimulus local direction of movement and speed are additional features driving the spatial integration involved in the perception of shapes and contours.

The simulations performed in this section will deal with the geometric connectivities constructed in Section \ref{sec:geometry}, which aim to model the neural connections in the cortical area V1 of the visual cortex \cite{BCCS}. These connectivities are defined in feature spaces which include the local orthogonal velocities, which are detected by V1 specialized cells.

\vspace{-8pt}
\subsection{Grouping with spatial features}
\vspace{-2pt}

The perceptual bias towards collinear stimuli has classically been associated to the long-range horizontal connections linking cells in V1 having similar preferences in stimulus orientation. This specialized form of intra-striate connectivity pattern is found across many species, including cats \cite{Kisvarday1997}, tree shrews \cite{Chisum2003}, and primates \cite{Angelucci2002}, the main difference being the specificity and the spatial extent of the connections. Furthermore, axons seem to follow the retinotopic cortical map anisotropically, with the axis of anisotropy being related to the orientation tuning of the originating cell \cite{Bosking1997}. 
The clustering algorithm with the affinity matrix $A_3^{H,\kappa}$ constructed as in (\ref{eq:corticalaff}) using the kernel $\hat\Gamma_3^{H,\kappa}$ may then be neurally motivated by the assumption that spatial integration, grouping and shape perception are fundamentally modulated by the position and the orientation of the elements in the visual space.
Other prominent features of the visual stimulus, namely its velocity and direction of motion, seem to play an important role in the spatial integration of oriented elements \cite{Rainville2005}.
In addition to stimulus orientation, cells in the striate areas are also selective for the direction of motion orthogonal to the cell's preferred orientation \cite{DeAngelis1995, BCS2013}.
These selectivities are also structurally mapped in the cortical surface, with nearby neurons being tuned for similar motion direction \cite{Weliky1996}, and it has been shown that excitatory horizontal connections in the V1 of the ferret are strictly iso-direction-tuned \cite{Roerig1999}.
In order to model the grouping effect of these connectivities we will then make use of the affinity matrix $A_0^{H,\kappa,\alpha}$ constructed with the extended kernel $\hat\Gamma_0^{H,\kappa,\alpha}$.

\begin{figure}[h!]
\centering
\includegraphics[trim=4cm 1cm 4.5cm 1cm, clip=true,width=.78\textwidth]{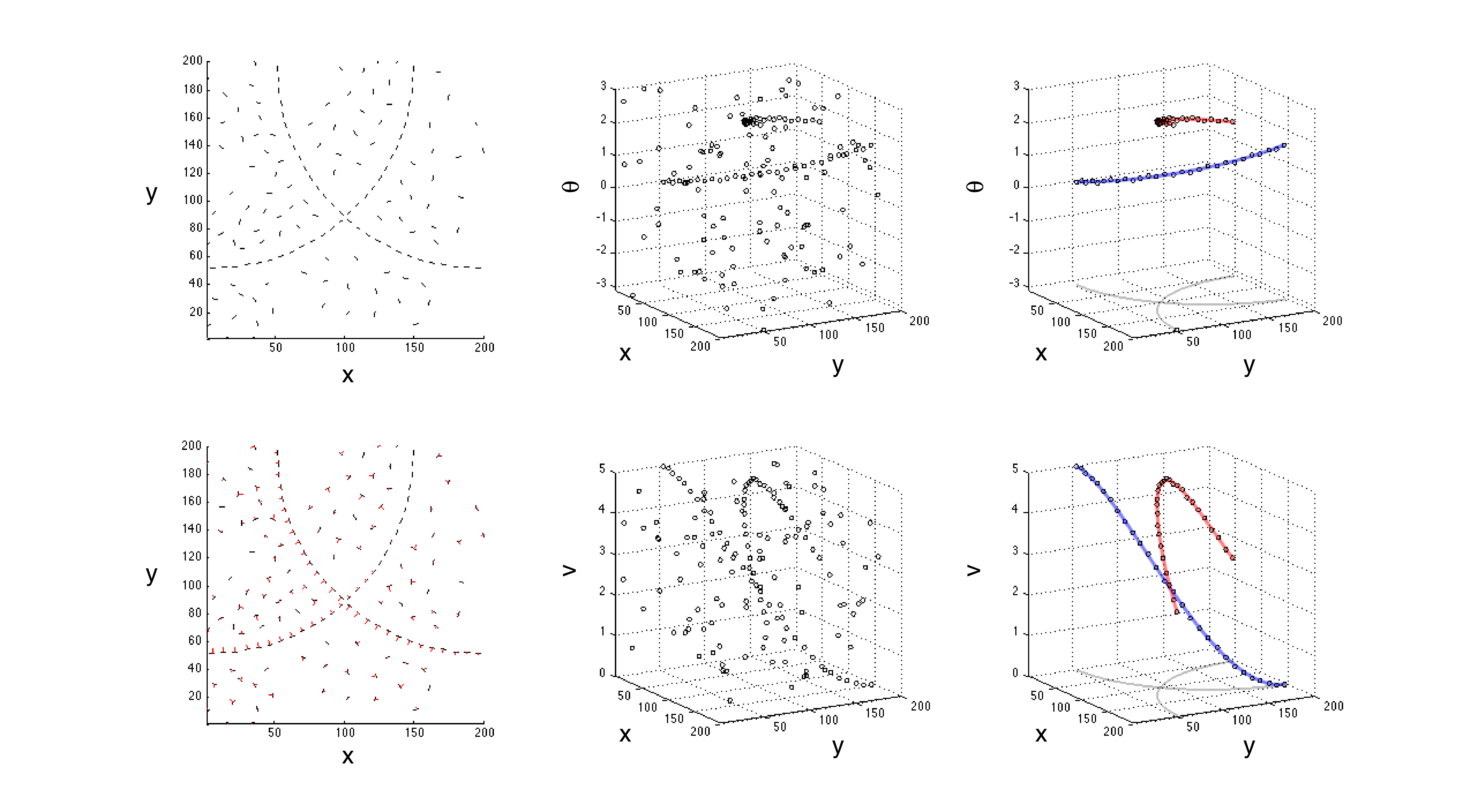}
\caption{An instance of the dataset $S^k_r$, where the left column shows the corresponding visual stimulus, the central column shows the dataset in the feature space, and the right column shows the perceptual units, which constitute the target segmentation objects, that for  this instance have curvature $k=0.023$. The top row shows the dataset in the feature space $\M_3 = \R^2 \times S^1$, while the bottom row shows it in the feature space $\M_0 = \R^2 \times S^1 \times \R^+$ with a velocity assignment producing a shape deformation. The bottom left image contains the spatial stimulus (black) with the magnitude of its istantaneous orthogonal velocity (red).}\label{fig:groupstim1}
\end{figure}

\vspace{-8pt}
\paragraph{The stimuli.}
We have considered a dataset made of $n$ points $\{\zeta_i\}_{i = 1}^n$ in the feature space of positions and orientations $\M_3 = \R^2 \times S^1$, having the following structure. Two perceptual units consisting of segments aligned along circle arcs with curvature $k$ are embedded in a background environment of $r$ segment having random positions and orientations. Such dataset will be denoted as $S^k_r$, and is depicted in Fig. \ref{fig:groupstim1}, top.

By assigning to each point $\zeta_i$ an instantaneous orthogonal velocity $v_i$, such dataset can be considered as $n$ points $\{\xi_i = (\zeta_i,v_i)\}_{i = 1}^n$ embedded in the larger feature space $\M_0 = \R^2 \times S^1 \times \R^+$. A velocity field that is constant on each perceptual unit would describe a rigid motion, and the analysis of such a case with the affinity $A_0^{H,\kappa,\alpha}$ can be understood as an extension of the static case, clearly providing grouping improvement.
\newpage
A more interesting case is that of a velocity field that changes along each perceptual unit, which represents \emph{shape deformation}. We have then generated a distribution of the velocity feature in the following way: given a fixed maximal velocity $V$, we have assigned to points belonging to the perceptual units a velocity feature $v_i$ that varies sinusoidally along the arc circle, passing from zero to $V$, and assigned a random velocity between $0$ and $V$ to the background points. Such dataset $\{\xi_i\}_{i = 1}^n \subset \M_0$ is depicted in Fig. \ref{fig:groupstim1}, bottom.

\vspace{-20pt}
\paragraph{The numerical experiments.}
We have applied the method described Section \ref{sec:cdm}, using the same spectral clustering parameters used for the examples of Fig. \ref{fig:sampleclus3}, i.e. $\epsilon = 0.05$, $\tau = 150$ and $M = 3$, on a set of stimuli $S^k_r$ for different values of $k$ and $r$.
We have used both the affinity $A_3^{H,\kappa}$ on $\M_0 = \R^2 \times S^1$, and $A_0^{H,\kappa,\alpha}$ on $\M_0 = \R^2 \times S^1 \times \R^+$, in this case setting the maximum velocity assignment to $V = 5$. We have then evaluated the grouping performances for various kernel parameter sets $(H,\kappa)$ and $(H,\kappa,\alpha)$ by computing, for each iteration, a percentage error measure
$$
E = \frac{ E_1 + E_2 + E_3 }{n}
$$
where $n$ is the total number of points in the stimulus, $E_1$ is the number of points that were incorrectly assigned to the noise/background set, $E_2$ is the number of random points that were incorrectly recognized as part of a perceptual unit, and $E_3$ is the points pertaining to an over- or under-partitioned contour. In order to correctly compare results obtained by partitioning the dataset with a different number of random points, we have then averaged $E$ over $100$ repetitions, where at each repetition we have changed the random part of the stimulus and calculated new kernels $\hat \Gamma_3^{H,\kappa}$ and $\hat \Gamma_0 ^{H,\kappa,\alpha}$. The resulting mean percentage error measure $\hat E$ has been finally taken as a measure of the quality of the grouping.

\begin{figure}[!h]
\centering
\includegraphics[trim=0cm 0cm 0cm 0cm, clip=true,width=.46\textwidth]{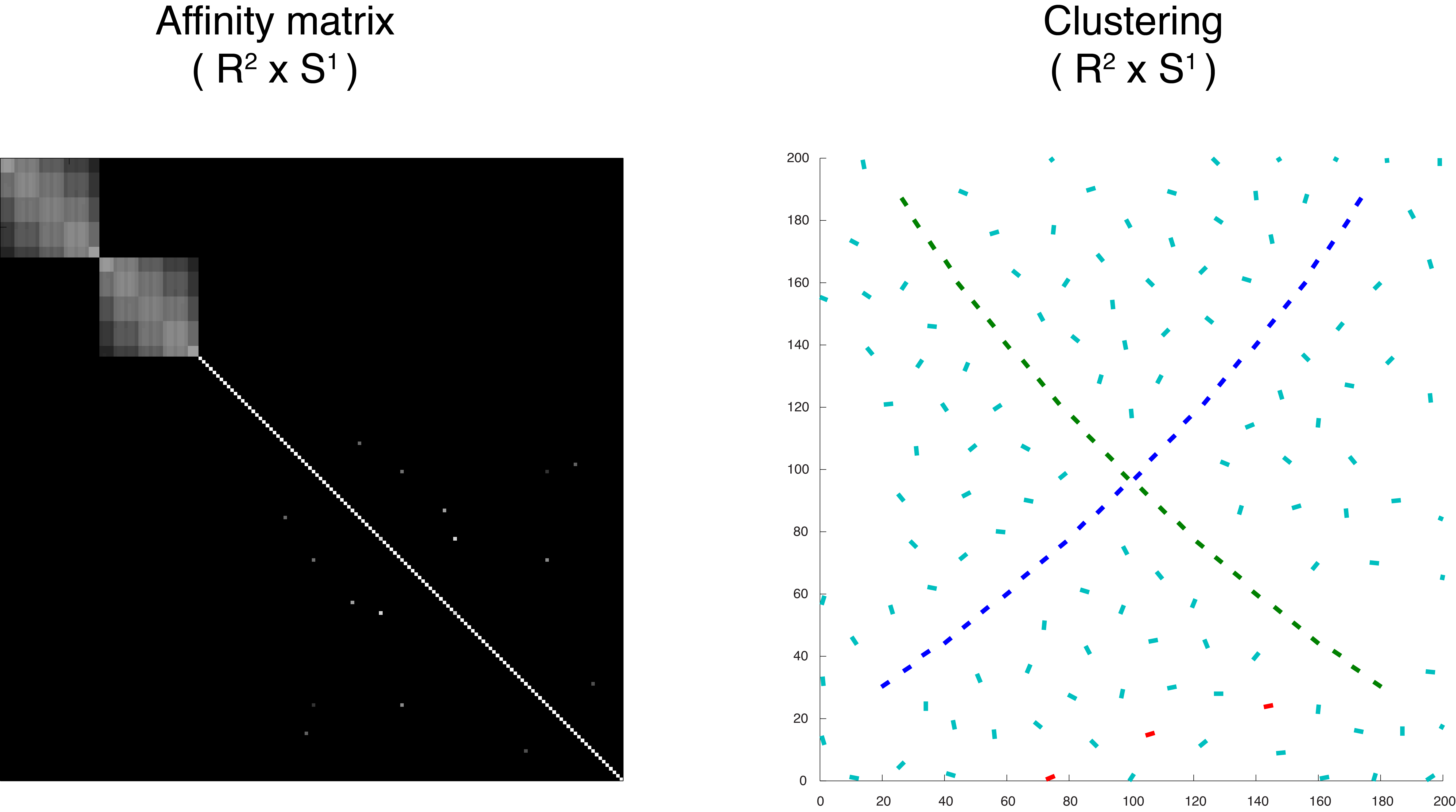}\hfill
\includegraphics[trim=0cm 0cm 0cm 0cm, clip=true,width=.46\textwidth]{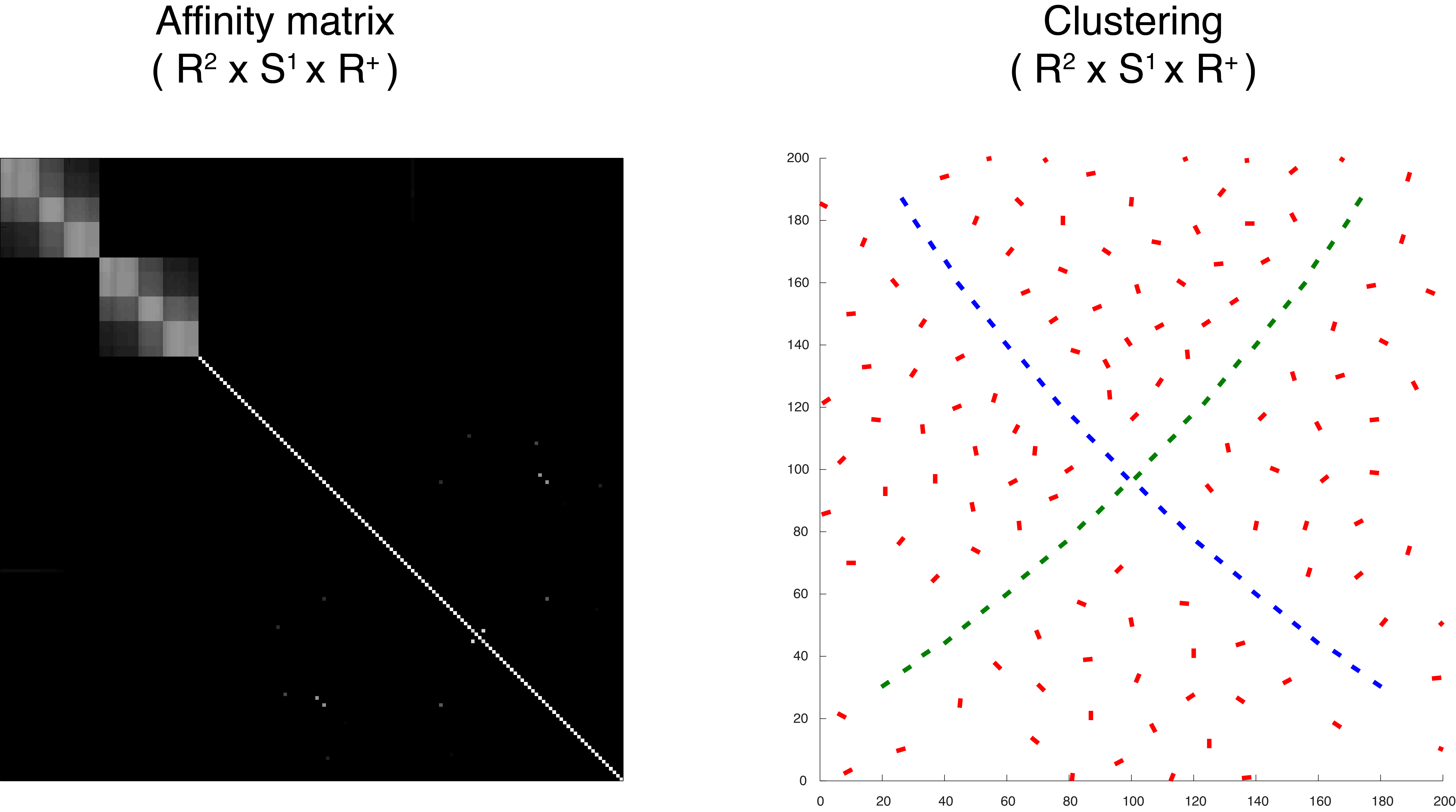}
\includegraphics[trim=0cm 0cm 0cm 0cm, clip=true,width=.46\textwidth]{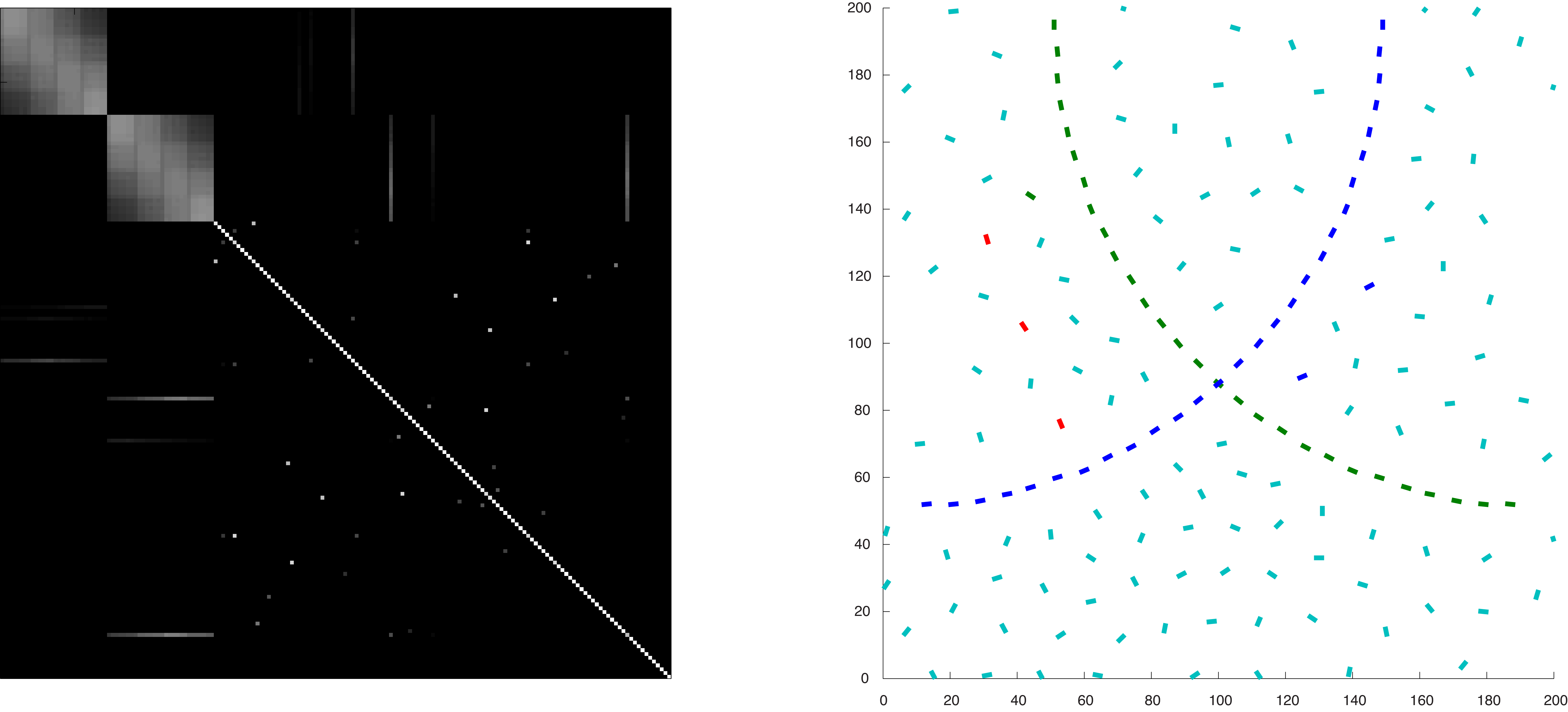}\hfill
\includegraphics[trim=0cm 0cm 0cm 0cm, clip=true,width=.46\textwidth]{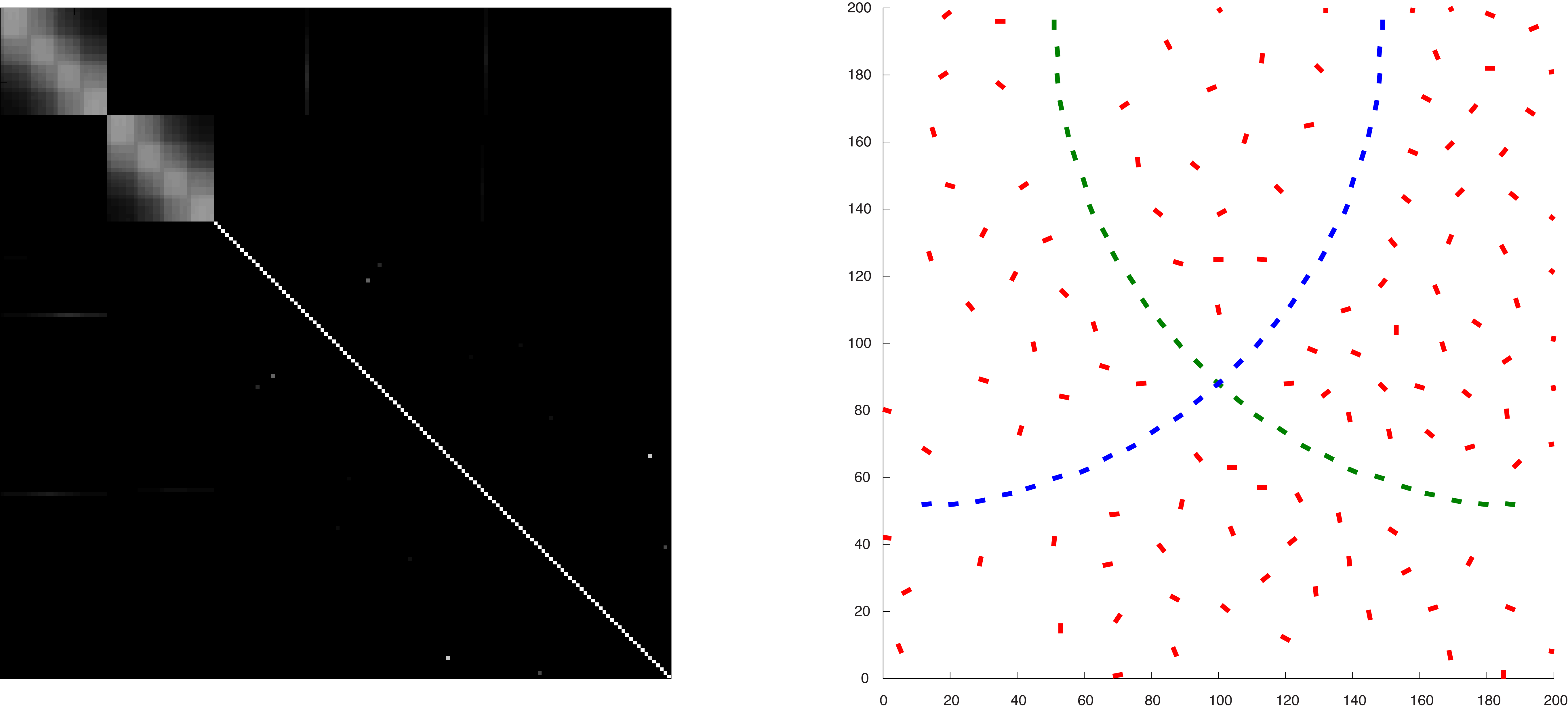}
\includegraphics[trim=0cm 0cm 0cm 0cm, clip=true,width=.46\textwidth]{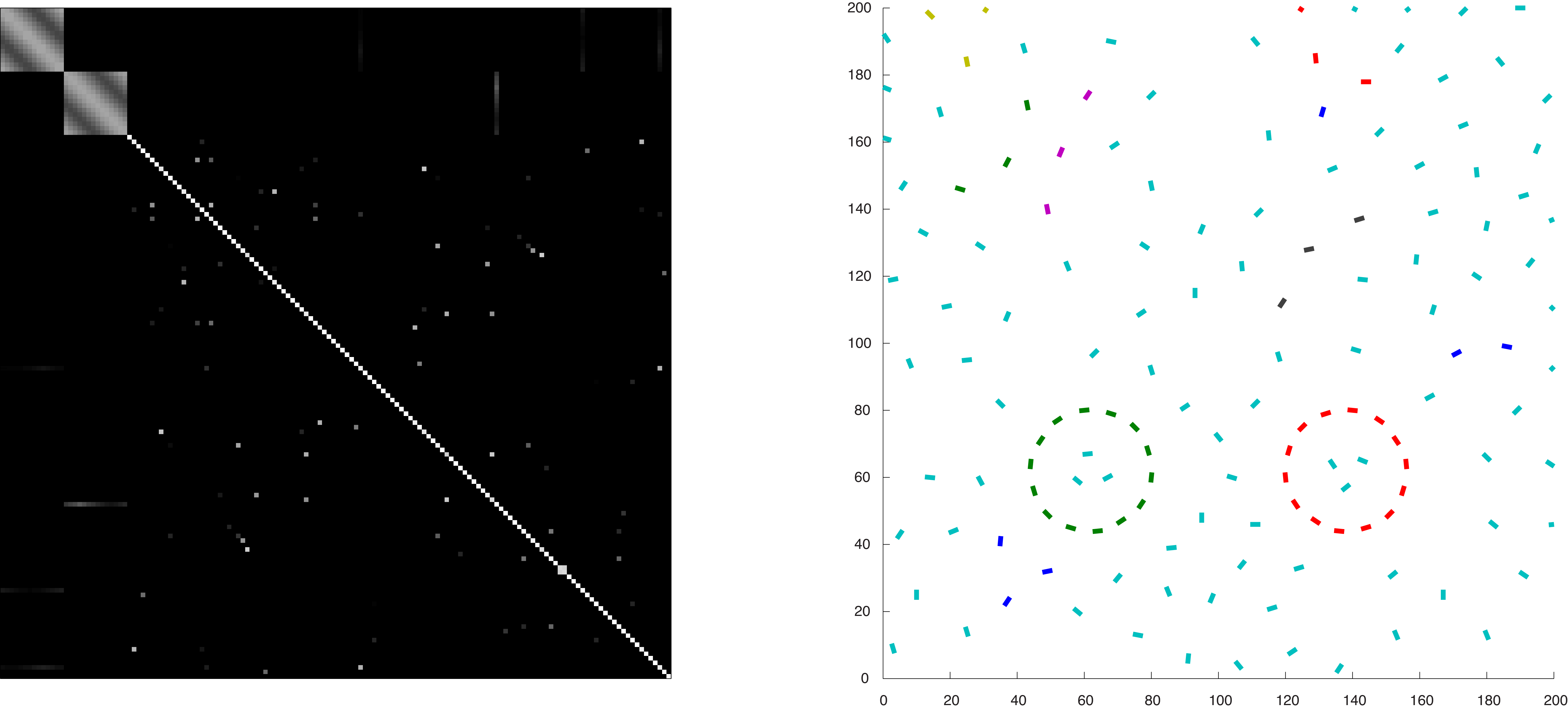}\hfill
\includegraphics[trim=0cm 0cm 0cm 0cm, clip=true,width=.46\textwidth]{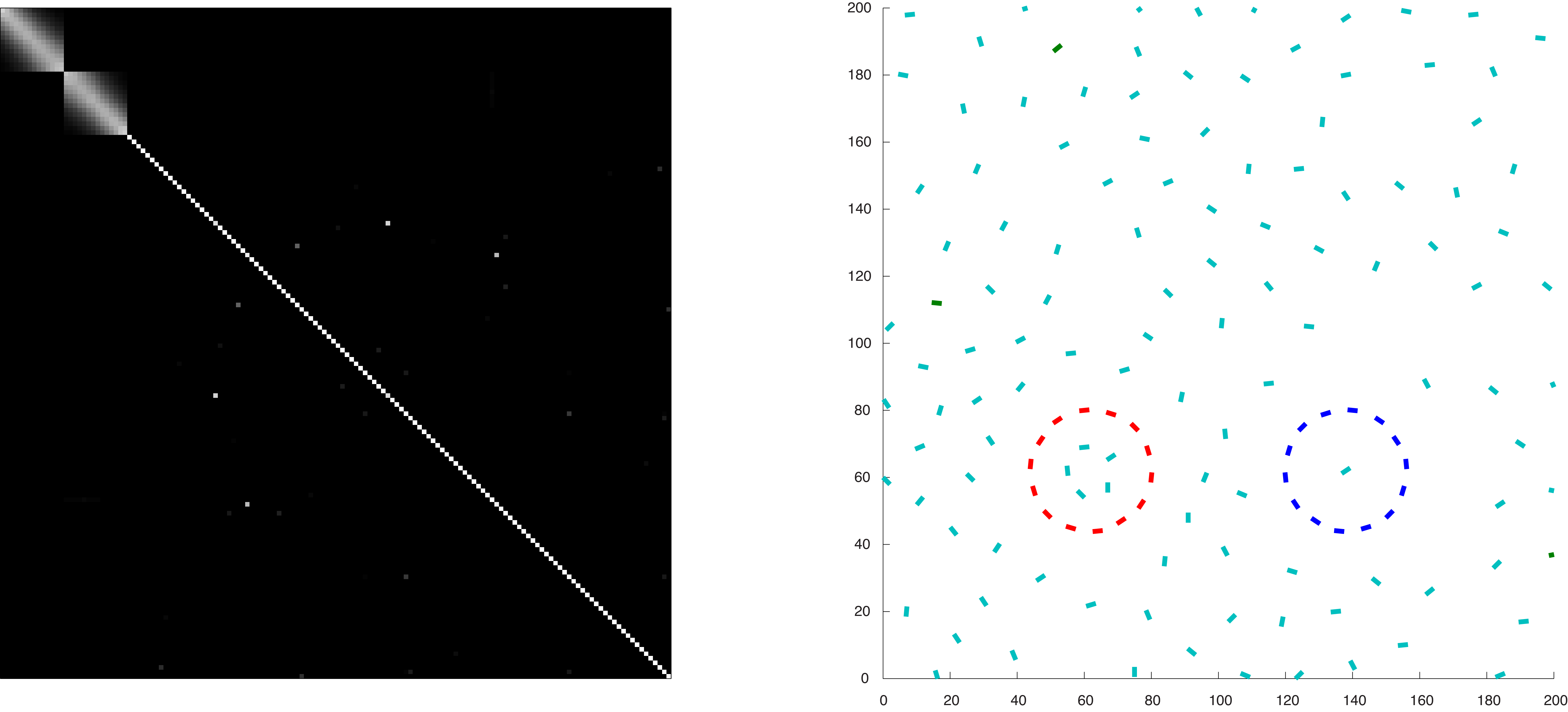}
\caption{Results of the proposed algorithm for data set with perceptual units with different curvatures. Left column: results obtained by using the position-orientation connectivity given by $\Gamma_3$. Right column: results obtained by using the position-orientation-velocity connectivity given by $\Gamma_0$. Affinity matrices built with $\Gamma_0$ are cleaner than those built with $\Gamma_3$: they generally avoid spurious affinities between perceptual units and noise, while maintaining the approximate object block diagonal structure.
}\label{fig:sampleclus4}
\end{figure}

\vspace{-8pt}
\paragraph{The results.}
\begin{figure}
\centering
\includegraphics[clip=true,width=.90\textwidth]{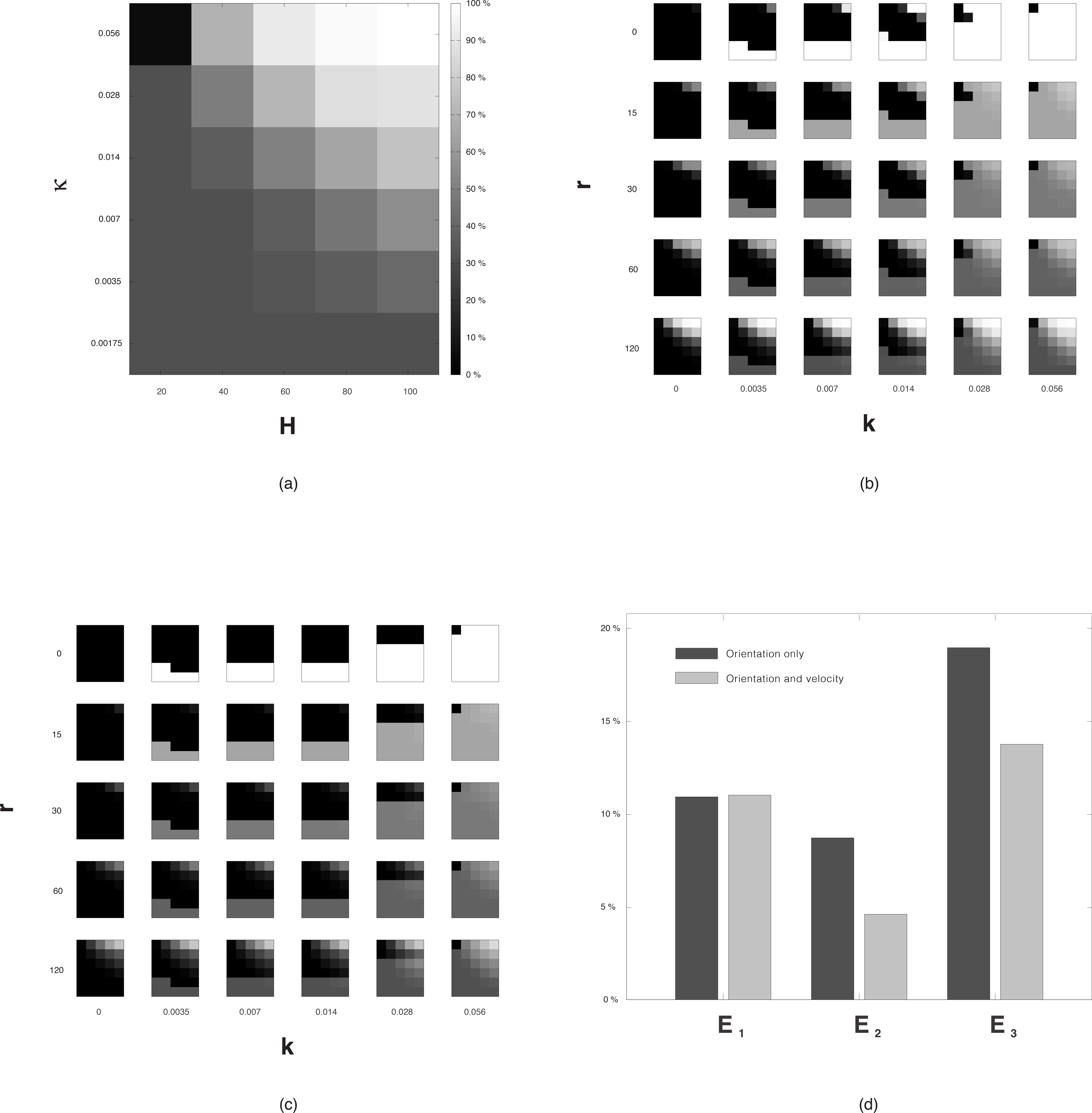}
\caption{Parametric analysis for visual grouping in $\R^2 \times S^1$ and $\R^2 \times S^1 \times \R^+$. The color intensity of each region is proportional to the percentage of misinterpreted points averaged over $100$ repetitions. (a) Grouping results for the stimulus $S^{0.056}_{120}$ in $\R^2 \times S^1$, by varying the kernel parameters $(\kappa,H)$. (b) Grouping results as in (a) for the stimuli $S^{k}_{r}$ in $\R^2 \times S^1$ by varying the stimulus parameters $(k,r)$. (c) Grouping results for the stimuli $S^{k}_{r}$ in $\R^2 \times S^1 \times \R^+$ with nonuniform velocity assignments. (d) Comparison of the grouping performances in position-orientation vs position-orientation-velocity with separated analysis of the three error sources.}\label{fig:grouping_res_ov}
\end{figure}
The results of the experiments are shown in Fig. \ref{fig:grouping_res_ov} and Fig. \ref{fig:sampleclus4}.

In Fig. \ref{fig:grouping_res_ov} $(a)$ we have represented with a grayscale intensity the error $\hat E$ for the spectral clustering of $S^{0.056}_{120}$ with the affinity $A_3^{H,\kappa}$ in $\R^2 \times S^1$ by varying the parameters $(H,\kappa)$ of the kernel. The parameters $(k,r) = (0.056,120)$ correspond to the stimulus in the feature space of positions and orientations composed of the perceptual units with the highest curvature, and having the highest number of random elements, among all the ones tested in the present analysis.
A first significative feature emerged is that the set of kernel parameters which gives the lowest grouping error value is $\kappa = 0.056, H = 20$, that is, the same curvature value of the contours in the stimulus and short stochastic path length.
Moreover, we observe that by maintaining the kernel parameter $H$ set to its minimum value and decreasing the kernel diffusion coefficient $\kappa$ we have had an approximately constant higher error value $\hat E$, whose dominating components are $E_1$ and $E_3$. This indicates that a reduction in the width of the fan of stochastic curves generating $\Gamma_3$ impairs the connectivity between high curvature contour elements, so that the algorithm perceives them as separate units and assigns part or all of them to the background/noise set.
It is also worth noting that, regardless of the kernel diffusion coefficient, increasing the parameter $H$ negatively impacts the quality of the grouping, this time mainly because of the error component $E_2$. This was predictable, as longest stochastic path lengths can generate affinities between elements very distant from each other. A high number of random elements can in this case induce the algorithm to recognize them as part of a distant contour. This gives its worse effects when both $\kappa$ and $H$ have high values, because with the associated kernel a point in $\M_3$ can potentially be connected to distant elements having a very different orientation. In such cases, also the error component $E_3$ gives its contribute, as the two perceptual units can be under-partitioned and interpreted as one unique object, having each one reciprocally affine contour elements.

Fig. \ref{fig:grouping_res_ov} $(b)$ resumes the same kind of grouping analysis carried out with $30$ different stimuli $S^k_r \in \M_3$ by varying the parameters $k$ and $r$. We can notice a positive correlation between the stimulus contour curvature $k$ and the smallest value of the kernel diffusion coefficient $\kappa$ that provides optimal grouping. Smaller values of $\kappa$ mostly lead to over-partitioned or unrecognized contours, while higher $\kappa$'s generally do not impair the contour grouping capability of the algorithm, even if we have observed that the error component $E_2$ tends to grow together with the number of random elements in the space $r$ and the length of the stochastic paths $H$.
In general thus, in order to correctly identify contours-like objects, the algorithm should be based on a connectivity whose diffusion coefficient is above the contour's curvature.

These results suggest that for a successful grouping the parameters $\kappa$ and $H$ could be associated to some property of the image detected by the visual cortex. In the discrete case for example, the model could be extended by adding two fiber variables, namely local stimulus curvature and scale, governing the numerical value taken by the two connectivity parameters. One could also argue - and our results are consistent with this view - that curvature and scale are very close concepts, strongly influencing each other, so that $\kappa$ and $H$ should not be independent. A deeper study of these aspects will be addressed in a future paper.
\newpage
In Fig. \ref{fig:grouping_res_ov} $(c)$ we show the grouping results obtained by analyzing the stimuli $S^k_r$ embedded in $\M_0$, with the described non uniform velocity assignment, using the connectivity $\hat\Gamma_0^{H,\kappa,\alpha}$. From a first visual inspection, we can see that the correlation constraint between the contour curvature $k$ and the orientation diffusion coefficient $\kappa$ is still present. However, the error at the highest values of $\kappa$ and $H$ is significantly reduced, if not almost completely eliminated for the stimuli having fewer random elements, due to the influence on the algorithm of correlations on the direction of motion.
We have not presented the detailed analysis on the parameter $\alpha$, which is very similar to the one related to $\kappa$. We limit ourselves to observe that the best results were obtained for values of $\alpha$ close to $\overline{\alpha} = \displaystyle\frac{\pi}{L}V$, where $L$ is the length of the perceptual unit (circle arc) and $V$ is the maximum velocity assigned, which was set to $V = 5$. The reason can be explained in terms of the velocity assignment, which as said we have chosen to be sinusoidally varying along the contour, namely
$$
v(\phi) = V \sin\left( \frac{\pi \phi}{L} \right) + \textrm{const.} 
$$
where $\phi \in [0,L]$ is the arc parameter of the circular curves. Then $\overline{\alpha} = \max \left | \dot v \right |$, which is a diffusion constant that ensures that grouping capabilities will be preserved even at the maximum local speed change rate along the contour. In particular, the results shown refer to clustering with the parameter $\alpha$ set to $\overline{\alpha}$.

A comparison of the grouping quality obtained with the orientation connectivity $\hat\Gamma_3^{H,\kappa}$ and with the the orientation and velocity connectivity $\hat\Gamma_0^{H,\kappa,\alpha}$, with separate presentation of the three error sources considere, is shown in Fig. \ref{fig:grouping_res_ov} $(d)$. The histograms correspond to the average of the three error components $E_1$, $E_2$ and $E_3$ over all the stimuli, the kernel parameters and the repetitions. The analysis of $E_1$ shows that both kernels tend to confuse the perceptual units with noise approximately in the same way. However, from the analysis of $E_2$ we can see that information on local velocity significantly enhances the performance of the algorithm in the presence of noise, improving the assignment of the random elements to the background. Also the analysis of $E_3$ shows that the connectivity $\hat\Gamma_0^{H,\kappa,\alpha}$ reduces the under-partitioning effect of the stimulus, which typically occurs when the parameters $(\kappa,H)$ make $\hat\Gamma_3^{H,\kappa}$ too long range and widespread.

Finally, some particular cases of the discussed grouping simulations are shown in Fig. \ref{fig:sampleclus4}, where one can see how the use of $\hat\Gamma_0^{H,\kappa,\alpha}$ instead of $\hat\Gamma_3^{H,\kappa}$ concurs in reducing different kinds of grouping errors of the algorithm.

\subsection{Spatio-temporal grouping}

In this subsection we consider grouping performed in space and time. Similarly to what happens for the integration of spatial visual information, our visual system is capable to easily predict stimulus trajectories, and to group together elements having similar motion or apparent motion paths. At the psychophysical level, the facilitation in detecting moving stimuli, given a previous cue with a coherent trajectory, is found to be significantly higher than the one expected from the temporal response summation given by the onset and offset dynamics of classical RPs \cite{Verghese1999}. One possible explanation could be the existence of a specialized facilitatory network linking cells anisotropically and coherently with their direction of motion.

In \cite{Ledgeway2002}, the perception of spatial contours defined by non-oriented stimuli moving coherently and tangentially along a path was studied. The authors suggested rules similar to the ones driving facilitation in position/orientation and inferred a possible role played by a trajectory-specialized network. Another possible evidence of a trajectory-driven connectivity comes from a recent study of the dynamics of neural population response to sudden change of motion direction, where it is shown that for low angular changes a non-linear part of the response provides a sort of spatio-temporal interpolation \cite{Wu2011}.

We recall also that, at the physiological level, although the basic mechanisms of velocity detection are already present in V1, the estimation of stimulus motion has been classically associated with neurons in visual area MT/V5. There, cells with high selectivity in direction of movement and extended sensitivities to a wide range of stimulus velocities are indeed present \cite{Maunsell1983, Rodman1987}. Extra-striate areas are retinotopically organized, with anisotropic and asymmetric connectivity bundles reaching columns of cells tuned for similar orientation and direction preference \cite{Malach1997}.

On the other hand, the important role played by V1 collinear horizontal connectivity in motion perception has been suggested already in \cite{Series2002}, after showing that perception of speed is biased by the direction of motion. 
Also, recent results in \cite{Pavan2011} showed how trajectories of oriented segments are significantly more detectable for orientations orthogonal to the path of motion, thus supporting the hypothesis of two different facilitatory mechanisms \cite{Hess2003}.

\begin{figure}
\centering
\includegraphics[trim=0cm 0cm 0cm 0cm, clip=true,width=\textwidth]{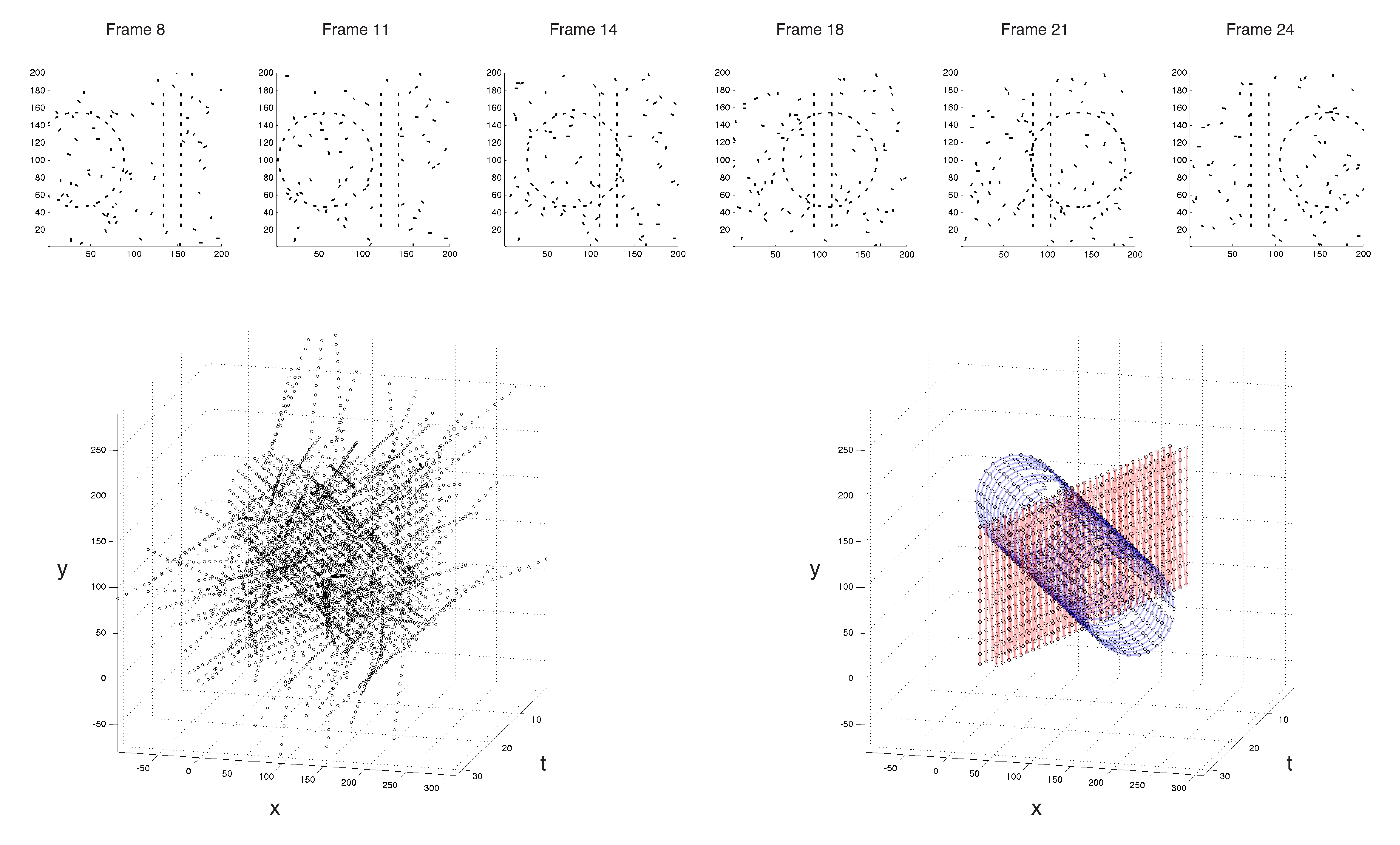}
\caption{An instantiation of the stimulus living in $\R^2 \times \R^+ \times S^1 \times \R^+$ used in the tests for spatio-temporal grouping of moving shapes. The circle has curvature $k=0.02$ and its velocity is $7.5$ units/frame, that is twice that of the bars ($3.75$ units/frame).}\label{fig:groupstim2}
\end{figure}

\paragraph{The stimulus.} The dataset that we have used in this simulation is depicted in Fig. \ref{fig:groupstim2}. We created a set of points $\eta_i = (x_i,y_i,t_i,\theta_i,v_i) \in \M_T = \R^2 \times \R^+ \times S^1 \times \R^+$, represented in the figures by segments moving in time for a total of $n_t = 32$ frames. Three perceptual units, a circular shape of curvature $k = 0.02$ and two bars, move as rigid bodies, with the circle translating with speed $V_{\textnormal{circ}} = 7.5$ spatial units/frame in the opposite direction to the bars, who both move with speed $V_{\textnormal{bars}} = 3.75$ spatial units/frame. We recall that the fiber coordinate of local velocities $v_i$ of each point represents just the projection onto $\vec X_3$, i.e. orthogonal to the segment orientation, of the real velocity vector. Similarly to what we did in the previous example, these perceptual units were embedded into a background consisting of a variable number $r$ of random elements, each one having a uniform motion path along their $\vec X_3$ direction during all the stimulus frames.

\paragraph{The numerical experiment.}
The aim of the grouping algorithm that we will use for this experiment is to carry out a segmentation of the full spatio-temporal surfaces representing the moving objects.

From the point of view of the structure of the stimulus which gives rise to the dataset, this represents a composite task. It consists of a visual grouping at the spatial level, which identifies the objects, and a visual grouping at the spatio-temporal level, where each previously identified cluster is recognized as constituting the same object during its movement. It is then reasonable to assume that the two connectivity mechanisms $\hat\Gamma_0^{H, \kappa, \alpha}$, modeling the interactions between points of a motion contour, and $\hat\Gamma_T^{H, \kappa, \alpha}$, which models motion integration of point trajectories, combine in the clustering of spatio-temporal perceptual units.

On the other hand, the presence of more than one grouping law governing the detection of contours, with different underlying implementing structures, was experimentally confirmed at the psychophysical level in \cite{Ledgeway2005}. Moreover, such composition of different mechanisms resulted to be compatibile with a probabilistic summation.

Guided by such arguments, we have then chosen to perform the spectral clustering on this dataset with a matrix $P$ obtained as the sum of the transition probability (normalized affinity) matrices obtained from the cortical connectivities $\hat\Gamma_0^{H, \kappa, \alpha}$ and $\hat\Gamma_T^{H, \kappa, \alpha}$. More precisely, we have constructed the symmetric matrix $A_0$, based on $\hat\Gamma_0^{H, \kappa, \alpha}$ as in (\ref{eq:corticalaff}), with the additional condition of setting zero affinity between points having different temporal coordinates, and we have constructed the nonsymmetric matrix $A_T$, based on $\hat\Gamma_0^{H, \kappa, \alpha}$ as in (\ref{eq:corticalaffasymm}). We have then normalized both of them as in (\ref{eq:normalizedaff}), obtaining the transition probabilities $P_0$ and $P_T$, and we have defined the combined spatio-temporal normalized affinity as
\begin{equation}\label{eq:Psum}
P = \frac{P_0 + P_T}{2}.
\end{equation}

From the neural point of view, we observe that by relying on the normalized affinity (\ref{eq:Psum}) we are implcitly assuming a much faster propagation along the connectivity defined by $\hat\Gamma_0^{H, \kappa, \alpha}$, with respect to the temporal dynamics. More precisely, by assigning zero $A_0$ affinity to temporally separated points of the dataset, we are considering spatial connections that fully operate at each single frame, which corresponds to an almost instantaneous velocity detection and a high horizontal transmission speed.

Finally, in order to perform the spectral clustering over $P$, we will use the approach described in Section \ref{sec:asymmetric}. Indeed,
since $A_T$ is not symmetric, $P$ will have in general complex eigenvalues and eigenvectors, but it was constructed in such a way to keep a probabilistic structure.


\begin{figure}
\centering
\includegraphics[trim=0cm 0cm 0cm 0cm, clip=true,width=.95\textwidth]{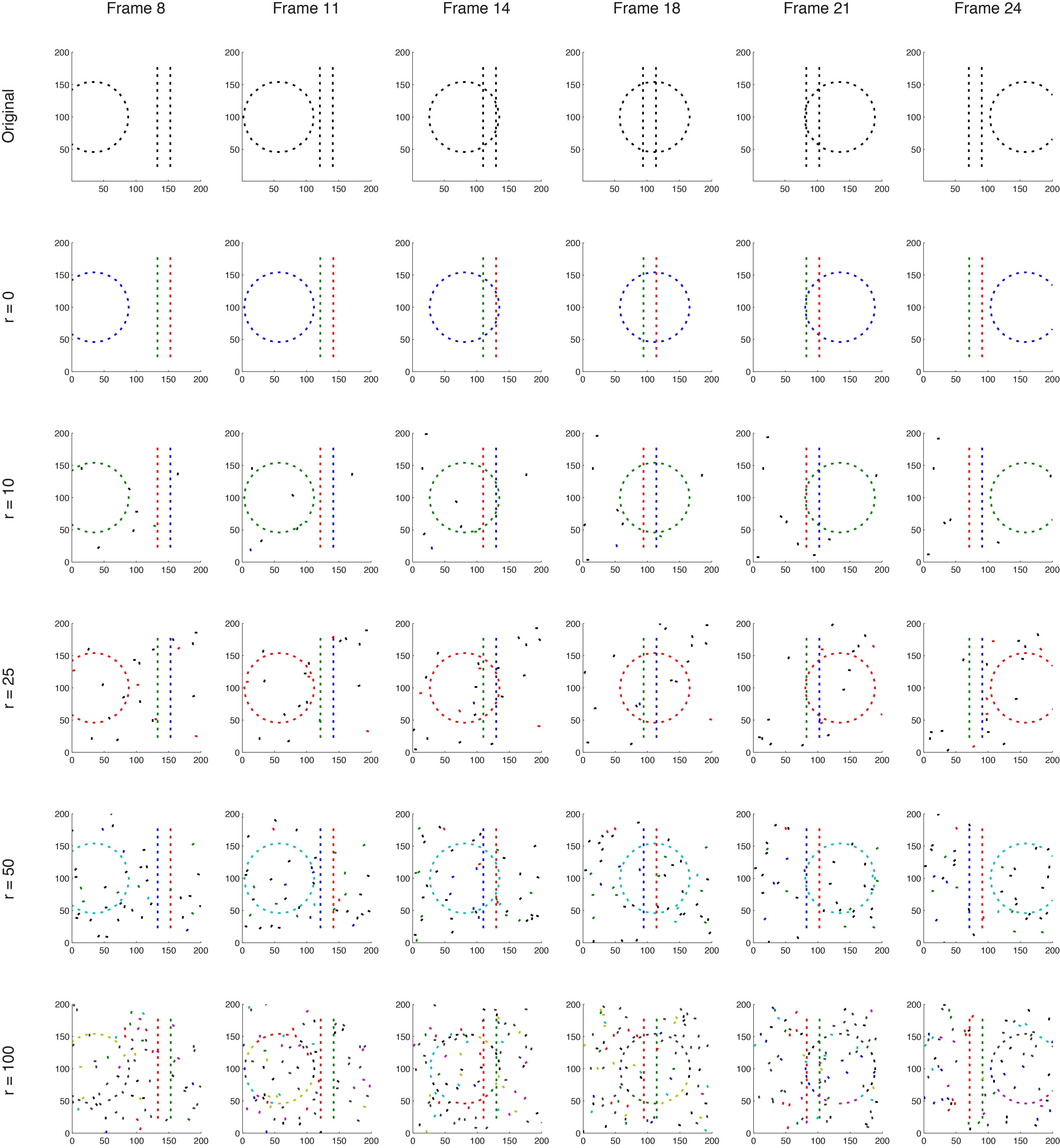}
\caption{Results obtained by using both connectivities $\hat\Gamma_0^{H, \kappa, \alpha}$ and $\hat\Gamma_T^{H, \kappa, \alpha}$ simultaneously. The grouping is succesful for stimuli with noise levels up to 50\%. At higher noise level the algorithm fails by over-partitioning the countours of the moving circle. The third row shows grouping results with $r = 50$ background elements, corresponding to about the 47\% of the total. The fourth row shows grouping results with $r = 100$ background elements, corresponding to about the 64\% of the total.}\label{fig:grouprestort}
\end{figure}

\paragraph{The results.}
The results obtained for various instances of the circle/bars stimulus by varying the number of background/noise elements $r$ are shown in Fig. \ref{fig:grouprestort}.

The parameters chosen to run the algorithm were $\epsilon = 0.01$, $\tau = 150$ and $M = 3$, where a smaller threshold $\epsilon$ with respect to the previous experiments is suggested by the presence of a semantically and geometrically sharper affinity. The integration parameter was set to $H = 40$ for both $\Gamma_0$ and $\Gamma_T$, according to the quality indications of the previous experiment. Similarly, we have set the angular diffusion parameters to the same value $\kappa = 0.014$ for both kernels.

With respect to the diffusion coefficient over velocities, we are showing the results for values that for both kernels stay close to the optimal value for the circle discussed in the previous subsection, that is $\ol{\alpha} = \frac{k V_{\textnormal{circ}}}{2} = 0.75$. Namely, we have set $\alpha = 0.5$ for $\Gamma_0$ and $\alpha = 1$ for $\Gamma_T$. In contrast to all other parameters, which were chosen to be equal for both kernels, this one is indeed observed to perform better when it is larger on the temporal connectivity. Such a behavior is coherent with the considerations made in \cite{Hess2003, Verghese2006}, where the effect of local changes in the velocity of a motion contour over the perception of visual units are studied. The results of these works show that such changes tend to weaken the visual grouping, but this effect is much stronger when changes are orthogonal to contours with respect to changes that are tangential to trajectories. In the present setting, we are performing visual grouping both at the level of motion contours, with $\Gamma_0$, and at the level of trajectories, with $\Gamma_T$, describing two copresent connectivities, and the parameter $\alpha$ describes the sensitivity of the corresponding kernel to local changes in the velocity, so the different levels of $\alpha$ reflect a higher sensitivity (lower diffusion) to velocity for the $\Gamma_0$ connectivity, and a lower sensitivity (higher diffusion) for the $\Gamma_T$ connectivity.

From Fig. \ref{fig:grouprestort} it is possible to see how spectral clustering with the composite affinity (\ref{eq:Psum}) performs in recognizing the spatio-temporal surfaces relative to the moving circle and the bars with different noise levels. It is succesful in clustering the perceptual units, separating clearly their boundaries from the background and between themselves, up to relativeli high noise levels. When the number $r$ of random elements was lower than about the 50\% of the total, we obtain correct clustering, but for higher noise values the algorithm began to give poor grouping results, as in the shown case of $r = 100$ random segments, which correspond to about 64\% of the total. It is worth noting that even at the higher noise values, the bars are always correctly retrieved. The algorithm tends indeed to fail in the detection of contours with high curvature, confusing them with the background segments, thus leading to over-partitioning.

In general, though, we show that the connectivity kernels defined by the proposed cortical-inspired geometrical model applied to a simple spectral clustering algorithm are able to carry out a non-trivial grouping task. To better understand the powerful mechanics involved in the calculations, let's consider for example that the only segments of the circle that present a positive affinity value with their corresponding points at future temporal positions are the ones having an orientation value near $\pm \frac{\pi}{2}$, as only for them the vector field $\vec X_5$ of connectivity propagation has the same direction of the global movement of the shape.

In fact, while the ability and the reliability of visual neurons in areas V1 and MT/V5 in measuring local stimulus orientation and speed have been studied extensively, the majority of cells in those areas respond solely to the local characteristics they are tuned for. In doing so, the measurements available in the first stages of the visual cortex are subject to the well-known aperture problem: with no information other than the local direction of movement, it cannot be said much about the real direction and speed of the object to which that local measurement refers. For a continuously moving contour, for example, classical orientation- and direction-selective cortical cells measure, for each position along the contour, just the velocity component that is orthogonal to the contour tangent direction at that point.
In the framework presented throughout the paper, this is modeled so that the fiber variable of local velocity refers to movements in the $\vec X_3$ direction.

\section{Conclusions}
\label{sec:conclusions}

In this work we have constructed a clustering algorithm for visual grouping of spatio-temporal stimuli in terms of geometric connectivity kernels associated to the functional architecture of the visual cortex. The main purpose was to test the segmentation capabilities of such a geometric model of low-level vision areas, with respect to spectral analysis mechanisms. Previous experimental investigations such as \cite{Ledgeway2005} already indicated that spatio-temporal perceptual associations play an important role in visual recognition tasks. The recent results in \cite{SartiCittiPercepts} also suggest the presence of concrete cortical implementations of a spectral analysis associated to lateral connection.

We have used recent dimensionality reduction methods \cite{Kannan}, chosen for their robustness and for their relatively simple structure, which allowed us to focus primarily on kernel properties. We have then performed several spectral clusterings on the introduced spatio-temporal cortical feature space of position, time, orientation and velocity, by using anisotropic affinities obtained by the models of cortical connections. Such affinities present a structure that is geometrically adapted to the considered stimuli, and proved to be able to better extract the relevant information compared to more classical kernels such as the isotropic Gaussian one. In particular, the proposed algorithm is capable to group together elements belonging to a single contour or shape moving in time, forming a spatio-temporal surface, and to distinguish them from a noisy background.

The first analysis that we carried out considered visual grouping when the affinity between points of a visual stimulus is assigned in the cortical feature space of positions and local orientations $\R^2 \times S^1$, and in the feature space $\R^2 \times S^1 \times \R^+$ taking locally detected velocities into account. The produced algorithm showed generally higher segmentation capabilities with respect to isotropic affinities defined only on the visual stimulus. Moreover, performance analyses showed that when using the additional velocity feature the results are less affected by the influence of random elements and proved significantly lower percentage of grouping errors due to over- or under-partitioning, thus allowing the horizontal connectivity to be spatially extended without suffering noise, as it happens in the visual cortex \cite{Bosking1997}. 


A second analysis considered grouping in space and time, and produced an algorithm that is able to identify spatial perceptual units and to follow them during their motion. This is done by extending the previous approach, combining the affinity over $\R^2 \times S^1 \times \R^+$, here treated as an instantaneous connectivity, with an affinity in the cortical feature space of positions, activation times, local orientations and locally detected velocities $\R^2 \times \R^+ \times S^1 \times \R^+$. In order to cope with the intrinsic causality of time evolution, we have worked direcly on asymmetric affinities and used probabilistic clustering arguments \cite{Pentney2005}. The copresence of more than one connectivity mechanism in the segmentation of spatio-temporal stimuli is actually a realistic assumption on visual cortex behavior \cite{Hess2003}.

We would like to remark that modelling the neural dynamical aspects of visual perception is a very delicate task. Indeed, many open questions remain to be addressed, regarding in particular how mean-field equations in space-time can be compatible with the fast time scales of visual processing, or how the delays introduced by the neural dynamics influence the functionalities of cortical architectures.
This first model of functional architecture in space-time does not take into account the integration time constants for neurons, as well as other biophysical phenemona, which certainly matter and deserve future study. These aspects are also strictly related to the general problem of understanding the time scale of mean-field dynamics at the physical level.
We observe however that this model of connectivity is able to implement a preactivation mechanism induced by spatio-temporal stimuli, whose biological plausibility was discussed and compared with neurophisiological measurement in \cite{BCCS}, showing good accordance.

The present investigation has thus provided a geometric framework to perform clustering in feature spaces, following principles of extensions that are feasible to be further generalized to higher dimensional detected features. Possible extensions could be achieved by the inclusion of features such as color, three dimensional stereo, or scale as additionaly detected features. Depending on the geometry of the stimuli, and on the psychophysiological indications, such extensions can lead to the definition of connectivities of higher complexity, or to the addition of copresent association mechanisms, following the two modalities that were discussed. This approach provides both a way to desgin artificial perceptual algorithms adapted to the geometry of the information, and a tool to propose new models of actual connections in the visual cortex, also suggesting further psychophysical or physiological experiments to compare and tune the model's parameters in order to fit real visual perception and cognition behaviors. Moreover, a natural future step will be the inclusion of realistic feature detection mechanisms allowing to apply this methodology to real stimuli.

\vspace{10pt}

\noindent
{\bf Giacomo Cocci}\\
DEI - Department of Electrical, Electronic, and Information Engineering ``Guglielmo Marconi'', University of Bologna. Viale del Risorgimento 2, 40136 Bologna, Italy.\\
E-mail: giacomo.cocci2@unibo.it

\noindent
{\bf Davide Barbieri}\\
Department of Mathematics, Autonomous University of Madrid. Facultad de Ciencias, Campus Cantoblanco, 28049 Madrid, Spain.\\
E-mail: davide.barbieri@uam.es

\noindent
{\bf Giovanna Citti}\\
Department of Mathematics, University of Bologna. Piazza di Porta San Donato 5, 40126 Bologna, Italy.\\
E-mail: giovanna.citti@unibo.it

\noindent
{\bf Alessandro Sarti}\\
CAMS - Centre d'Analyse et de Math\'{e}matique Sociales. EHESS, 190-198 Avenue de France, 75244 Paris, France.\\
E-mail: alessandro.sarti@ehess.fr

\end{document}